\documentclass{article}

\usepackage[nonatbib]{neurips_data_2024}

\usepackage[utf8]{inputenc} % allow utf-8 input
\usepackage[T1]{fontenc}    % use 8-bit T1 fonts
\usepackage{hyperref}       % hyperlinks
\usepackage{url}            % simple URL typesetting
\usepackage{booktabs}       % professional-quality tables
\usepackage{amsfonts}       % blackboard math symbols
\usepackage{nicefrac}       % compact symbols for 1/2, etc.
\usepackage{microtype}      % microtypography
\usepackage{xcolor}         % colors
\usepackage{enumitem}
\usepackage{amsmath} % for using \mathcal
\usepackage{caption}
\usepackage{booktabs}
\usepackage{multirow}
\usepackage{graphicx}
\usepackage{amssymb}

\usepackage{tikz}
\usepackage{etoolbox}
\newcommand{\circled}[2][]{\tikz[baseline=(char.base)]
    {\node[shape = circle, draw, inner sep = 0.7pt]
    (char) {\phantom{\ifblank{#1}{#2}{#1}}};%
    \node at (char.center) {\makebox[0pt][c]{#2}};}}
\robustify{\circled}

\title{\texttt{GraphFM}: A Comprehensive Benchmark for Graph Foundation Model}

\author{%
Yuhao~Xu$^{1}$ \quad Xinqi~Liu$^{1}$ \quad Keyu~Duan$^2$ \quad Yi~Fang$^{1}$ \quad Yu-Neng~Chuang$^3$ \\ {\bf Daochen~Zha}$^3$ \quad {\bf Qiaoyu~Tan}$^{1}$ \\
$^1$Department of Computer Science, New York University Shanghai \\ $^2$Department of Computer Science, National University of Singapore \\ $^3$Department of Computer Science, Rice University \\
\texttt{\{yx3534, xl4600, yf2722, qiaoyu.tan\}@nyu.edu} \\
\texttt{k.duan@u.nus.edu}, \texttt{\{ynchuang, daochen.zha\}@rice.edu}
}

\begin{document}

\maketitle

\begin{abstract}
Foundation Models (FMs) serve as a general class for the development of artificial intelligence systems, offering broad potential for generalization across a spectrum of downstream tasks. Despite extensive research into self-supervised learning as the cornerstone of FMs, several outstanding issues persist in Graph Foundation Models that rely on graph self-supervised learning, namely: 1) \textbf{Homogenization}. The extent of generalization capability on downstream tasks remains unclear. 2) \textbf{Scalability}. It is unknown how effectively these models can scale to large datasets. 3) \textbf{Efficiency}. The training time and memory usage of these models require evaluation. 4) \textbf{Training Stop Criteria}. Determining the optimal stopping strategy for pre-training across multiple tasks to maximize performance on downstream tasks. To address these questions, we have constructed a rigorous benchmark that thoroughly analyzes and studies the generalization and scalability of self-supervised Graph Neural Network (GNN) models. Regarding generalization, we have implemented and compared the performance of various self-supervised GNN models, trained to generate node representations, across tasks such as node classification, link prediction, and node clustering. For scalability, we have compared the performance of various models after training using full-batch and mini-batch strategies. Additionally, we have assessed the training efficiency of these models by conducting experiments to test their GPU memory usage and throughput. Through these experiments, we aim to provide insights to motivate future research. The code for this benchmark is publicly available at \href{https://github.com/NYUSHCS/GraphFM}{https://github.com/NYUSHCS/GraphFM}. 
  % \qiaoyu{We should add the github link of our code. That is, the submission is not anonymous. Add all authors! }
\end{abstract}

\section{Introduction}
Foundation Models (FMs) represent an emerging paradigm of AI, focused on pre-training models on large datasets and subsequently adapting them to various downstream tasks~\cite{bommasani2021opportunities}. FMs have already made significant strides in the field of Natural Language Processing (NLP), driven by the remarkable success of Large Language Models (LLMs)~\cite{devlin2018bert,brown2020language,radford2021learning,raffel2020exploring,lewis2019bart,zhou2023comprehensive}. Inspired by their success in NLP, FMs have naturally emerged as prominent research subjects across various other domains, such as computer vision~\cite{yuan2021florence,bai2023qwen}, time series analysis~\cite{zhou2023one,jin2023time}, and recommender systems~\cite{geng2023vip5}.

Graph learning is also evolving towards Graph FMs, propelled by advancements in Graph Self-Supervised Learning (GSSL)~\cite{liu2023towards,xie2022self,wu2021self}. In GSSL, models are trained by solving auxiliary tasks, using supervision signals derived directly from the data itself without the need for human annotations. Consequently, GSSL is an effective approach to realizing Graph FMs by pre-training graph models on large unlabeled graphs. Existing GSSL methods typically follow two paradigms: contrastive models and generative models. Contrastive models generate two graph views through data augmentation and employ graph neural networks (GNNs) to learn representations by optimizing a contrastive objective~\cite{zhu2021empirical}. Generative models parameterize the encoder using GNNs~\cite{kipf2016semi, tan2019deep} and train the model by reconstructing observed edges~\cite{kipf2016variational,pan2018adversarially,shi2023gigamae} or node attributes~\cite{wang2017mgae,meng2019co}.

%The Foundation Model is an emerging paradigm for building AI systems based on a class of general-purpose models~\cite{bommasani2021opportunities,}. Foundation Models are trained on large datasets to create pre-trained models that can be widely adapted to downstream tasks, such as BERT~\cite{devlin2018bert}, GPT~\cite{brown2020language}, and CLIP~\cite{radford2021learning}. The Foundation Model~\cite{devlin2018bert, radford2019language,liu2019roberta,raffel2020exploring,lewis2019bart,zhou2023comprehensive} has been extensively researched and applied in the fields of NLP and CV.

%Inspired by the success of these Foundation Models in their respective domains, the Graph Foundation Model for processing graph-structured data has also been proposed~\cite{liu2023towards}. Self-supervised learning has become the cornerstone of Foundation Models due to its superior scalability and generalization capabilities~\cite{bommasani2021opportunities}. Graph self-supervised learning (GSSL) typically follow two paradigms: contrastive models~\cite{zhu2021empirical} and generative models~\cite{kipf2016variational}. Contrastive models usually generate two graph views through data augmentation, learn representations via graph neural networks (GNNs), and optimize using a contrastive objective. Generative models often parameterize the encoder using GNNs~\cite{kipf2016semi, tan2019deep} and train the model by reconstructing observed edges~\cite{kipf2016variational,pan2018adversarially} or node attributes ~\cite{wang2017mgae,meng2019co}. 

However, despite the plethora of proposed GSSL methods, it remains unclear how much progress we have made towards Graph FMs. \emph{(i) There is no clear understanding of the homogenization~\cite{liu2023towards}, or generalization across different downstream tasks, of existing GSSL methods.} The majority of GSSL algorithms predominantly concentrate on node classification tasks, with limited evaluation on other downstream tasks~\cite{hou2022graphmae,hou2023graphmae2,bielak2022graph,tan2019deep,zhang2021canonical,thakoor2021large,yi2024gaugllm}. Conversely, some are exclusively tailored to address link prediction tasks \cite{zhang2018link} or clustering tasks \cite{liu2023dink}. Thus, there is a lack of evaluation to understand how each GSSL method performs on all tasks. \emph{(ii) Existing GSSL methods are evaluated under different settings, leading to results that are not directly comparable.} For example, S2GAE~\cite{tan2023s2gae} is evaluated by an SVM classifier to do node classification task, while GraphMAE~\cite{hou2022graphmae} uses MLP. For hyperparameters, CCA-SSG~\cite{zhang2021canonical} searches for the learning rate in [5$e$-4, 1$e$-3, 5$e$-3], while GrapMAE2~\cite{hou2023graphmae2} explores [2.5$e$-3, 2$e$-3, 1$e$-3]. Such critical details can have a substantial impact on performance, yet they are not thoroughly addressed in the existing literature. \emph{(iii) There is a deficiency in evaluating the performance of GSSL methods across datasets of varying scales using different sampling strategies.} Some methods have only been evaluated on small datasets, lacking experimental validation on large-scale data \cite{zhu2021graph,zhang2021canonical,bielak2022graph}, where full-batch training is often impractical, necessitating mini-batch training with specific sampling strategies. In this case, the selection of sampling strategies can significantly impact performance, underscoring the need for a more comprehensive evaluation.
\begin{figure}[t]
    \centering
    \includegraphics[scale=0.37]{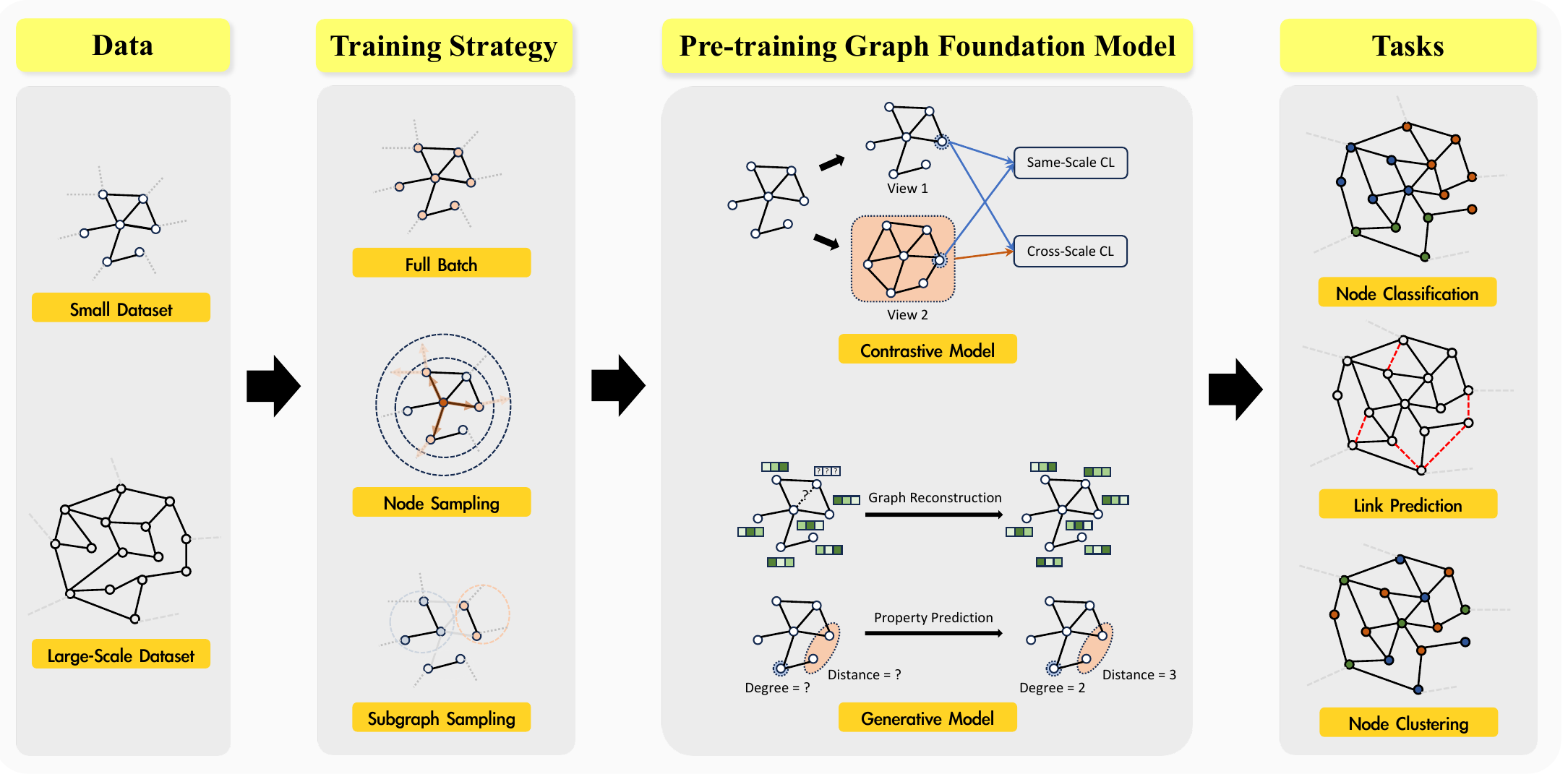}
    \caption{An overview of \texttt{GraphFM}. We perform a comprehensive benchmark of state-of-the-art self-supervised GNN models through four key aspects: dataset scale, training strategies, GSSL methods for Graph FMs, and adaptability to different downstream tasks.}
    \label{architecture}
    %The architecture of \texttt{GraphFM}. For small datasets, we use three training strategies to complete pre-training. For large-scale datasets, we employ node sampling and subgraph sampling methods for pre-training. For both types of datasets, the node representations obtained from the pre-trained models are used to perform three downstream tasks: node classification, link prediction, and node clustering.
\end{figure}

To bridge this gap, we introduce \texttt{GraphFM}, the first comprehensive benchmark for building Graph FMs based on GSSL. An overview of \texttt{GraphFM} is depicted in Figure~\ref{architecture}. \texttt{GraphFM} rigorously evaluates combinations across four key aspects: dataset scale, training strategies, various GSSL methods for Graph FMs, and adaptations to different downstream tasks. For a fair comparison, we implement all GSSL methods within a unified framework and employ consistent data processing and splitting methods for both training and evaluation. Additionally, we conduct hyperparameter searches with the same search budgets for all methods. In summary, our contributions include:

%To bridge this gap, we introduce \texttt{GraphFM}, the first comprehensive benchmark for Graph Foundation Models based on GSSL and the whole architecture of \texttt{GraphFM} is shown in Figure~\ref{architecture}. \texttt{GraphFM} compiles the most prominent GSSL models, including both contrastive and generative paradigms. It employs standardized data processing and data splitting methods and evaluates downstream tasks using a unified evaluation framework. By benchmarking \texttt{GraphFM} on various datasets, we make the following contributions:

\setlist[itemize]{leftmargin=12pt}
\begin{itemize}
\item {\bf Comprehensive benchmark.}~\texttt{GraphFM} enables a fair comparison among eight representative GSSL methods under a unified experimental setup across six popular datasets with varying scales. 
\item {\bf Multi-dimensional Analysis.}~\texttt{GraphFM} employs both full-batch and mini-batch training strategies and utilizes the trained node representations to perform three downstream tasks: node classification, link prediction, and node clustering. We systematically analyze the performance and efficiency under various settings. Furthermore, we investigate the influence of using performances from different downstream tasks or alternative metrics as early stopping criteria to train Graph FMs.
\item {\bf Openness.}~We have open-sourced our code and the pre-trained models on GitHub to facilitate future research. Based on our benchmark findings, we also outline potential future research directions to inspire further studies.
\end{itemize}

% The rest of the paper is structured as follows. In Section 2, we introduce the formulations and background of \texttt{GraphFM}. Section 3 details the design process of the benchmark. In Section 4, we analyze the training results and training efficiency of \texttt{GraphFM}.

\section{Preliminaries}
\label{sec:2}

\textbf{Notations and Problem Formulation.} Let $\mathcal{G} = (\mathcal{V}, \mathcal{E}, \mathbf {A}, \mathbf {X})$ be a graph, where $\mathcal{V}$ is the set of $N$ nodes, and $\mathbf {A} \in \mathbb{R}^{N \times N}$ is the adjacency matrix. $\mathcal{E}$ denotes the edge set and $\mathbf {X} \in \mathbb{R}^{N \times d}$ represents the corresponding feature matrix with dimension $d$. Typically, a GNN model is parameterized by a mapping function $f: (\mathbf {A}, \mathbf {X}) \rightarrow \mathbf{H} \in \mathbb{R}^{N \times l}$, which maps each node $v \in \mathcal{V}$ into a $l$-dimensional embedding vector $\mathbf{h}_v \in \mathbb{R}^l$, where $\mathbf{h}_v$ is the $v$-th row of $\mathbf{H}$. Once we obtain $\mathbf{H}$, we can adapt them with a head to perform downstream tasks. The objective of Graph FMs is to train a model that can generate high-quality $\mathbf{H}$, typically with GSSL methods, such that the adapted models can perform well across various downstream tasks.

%\daochen{Define notations, e.g., how to represent graph, nodes, links, etc? Based on the notation, what is the problem of the training graph foundation model? For example, we aim to pretrain XXX and achieve good performance on various downstream task, etc. Try to refer some other graph self-supervised learning papers and follow them to write formulations.}

{\bf Homogenization of Graph FMs.} Homogenization means the generalization capability of a FM to different downstream tasks~\cite{liu2023towards}. In the context of Graph FMs, we focus on three common tasks, including node classification, link prediction, and node clustering.

%{\bf Homogenization.} Homogenization~\cite{liu2023towards} means the generalization for Graph Foundation Model applied to different downstream tasks, such as node classification, link prediction and node clustering. To achieve homogenization of Graph Foundation Models, we need to decide on a unified form for different types of graph tasks. For example, datasets like collab and ppa from OGB~\cite{hu2020open} are not suitable for node classification tasks because they lack node labels.

{\bf Scalability and Training Strategies.} 
To train Graph FMs, it is often crucial to employ GSSL methods on large graphs. Standard GNNs typically operate in a full-batch setting, retaining the entire graph structure during forward and backward propagation to facilitate message passing (MP) among nodes. However, as the graph size increases, full-batch training becomes impractical due to significant memory usage and extensive computation time~\cite{duan2022comprehensive}. In this scenario, mini-batch training strategies can be adopted, using sampled subgraphs as mini-batches to approximate full-batch message passing, thereby significantly reducing memory consumption. Specifically, MP with $K$ layers can be expressed as follows:

\[
\mathbf{X}^{(K)} = \mathbf{A}^{(K-1)} \sigma \left( \mathbf{A}^{(K-2)} \sigma \left( \cdots \sigma \left( \mathbf{A}^{(0)} \mathbf{X}^{(0)} \mathbf{W}^{(0)} \right) \cdots \right) \mathbf{W}^{(K-2)} \right) \mathbf{W}^{(K-1)}
\]

where $\sigma$ is an activation function (e.g. ReLU) and $\mathbf{A}^{(l)}$ is the weighted adjacency matrix at the $l$-th layer. In the full-batch setting, $\mathbf{A}^{(l)}$ encompasses all nodes in the graph, while in the mini-batch setting, $\mathbf{A}^{(l)}$ only covers a subset of the nodes, resulting in $\mathbf{A}^{(l)}$ being a sub-matrix of the full adjacency matrix. The choice of sampling strategy plays an important role; in this work, we focus on two commonly used sampling strategies: node sampling \cite{hamilton2017inductive} and subgraph sampling \cite{chiang2019cluster}.

%{\bf Training Strategies and Scalability.} Scalability means the training performance of Graph Foundation Model in the large-scale dataset. As MP requires nodes to aggregate information from their neighbors, the entire graph structures must be maintained during forward and backward propagation, resulting in significant memory usage and extended computation time~\cite{duan2022comprehensive}. Many methods have been proposed to address this problem, such as GraphSAGE~\cite{hamilton2017inductive} and Cluster-GCN~\cite{chiang2019cluster}. These methods fall under the category of sampling-based methods. They perform batch training by utilizing sampled subgraphs as mini-batches to approximate full-batch message passing, thereby significantly reducing memory consumption. The model of \texttt{GraphFM} is based on the MP strategy, a unified formulation of MP with $K$ layers is presented as follows:
%\[
%\mathbf{X}^{(K)} = \mathbf{A}^{(K-1)} \sigma \left( \mathbf{A}^{(K-2)} \sigma \left( \cdots \sigma \left( \mathbf{A}^{(0)} \mathbf{X}^{(0)} \mathbf{W}^{(0)} \right) \cdots \right) \mathbf{W}^{(K-2)} \right) \mathbf{W}^{(K-1)}
%\]
%where $\sigma$ is an activation function (e.g. ReLU) and for full batch $\mathbf{A}^{(l)}$ is the weighted adjacency matrix at the $l$-th layer. For mini batch $\mathbf{A}^{(l)}$ will be replaced by $\widetilde{\mathbf{A}}_{\mathcal{B}_l}^{(k)}$, where $\mathcal{B}_l$ is the set of sampled nodes for the $l$-th layer, and $\widetilde{\mathbf{A}}^{(l)}$ is the adjacency matrix for the $l$-th layer sampled from the full graph..

{\bf Early stopping criteria.} When pre-training GNN models, we commonly employ early stopping and save the best model based on the performance of a specified metric on the validation set. Subsequently, we evaluate this saved model on the test set. This process is straightforward when focusing on a single task and evaluation metric, as often seen in the GSSL literature. However, 
% While this approach is straightforward when focusing on a single task and evaluation metric, 
% While this approach is straightforward when focusing on a single task and evaluation metric, 
training Graph FMs requires achieving good performance across various downstream tasks and metrics, such as accuracy and AUC. The impact of early stopping criteria on this objective has not been fully explored.

% {\bf Early stopping criteria.} During the training process of GSSL models, we need to save the pre-trained model that performs best on the validation set for testing on the test set. For conventional GSSL models, this is straightforward as they typically focus on a single task during training. However, in \texttt{GraphFM}, we aim to achieve high performance across various tasks. Saving pre-trained models based on different downstream tasks might impact performance on other tasks. In node classification tasks, accuracy is commonly used as the metric for saving pre-trained models. In link prediction tasks, AUC is typically used as the metric for saving pre-trained models\cite{tan2023s2gae}.
%\daochen{We often choose the best model based on validation performance. This process is straightforward for regular GNN training since we only have one task. However, in \texttt{GraphFM}s, we aim to achieve performance on various tasks...}

%Let $ \mathcal{G} = (\mathcal{V}, \mathcal{E}, {\mathbf{A}}, {\mathbf{X}}) $ be a graph, where $\mathcal{V}$ is the set of $\textit{N}$ nodes and $\mathbf{A} \in \mathbb{R}^{N \times N} $ is the adjacency matrix. $\mathcal{E}$ denotes the edge set and $\mathbf{X} \in \mathbb{R}^{N \times d}$ represents the corresponding feature matrix with dimension $d$. f : (\mathbf{A}, \mathbf{X}) \rightarrow \mathbf{H} \in \mathbb{R}^{N \times l}

\section{Benchmark Design}

We begin by introducing the datasets used in our benchmarking process, along with the algorithm implementations. Then, we pose the research questions to guide our benchmarking study.

\subsection{Dataset and Implementations}

{\bf Datasets.} To conduct a comprehensive evaluation of existing GSSL methods, we selected six widely used graph node classification datasets from the GSSL literature. Table~\ref{datasets} shows the statistical data of datasets, these datasets vary in size, allowing us to assess the generalization capabilities of current methods across different data scales. Specifically, we utilized three classic citation datasets: Cora, Citeseer, and Pubmed~\cite{sen2008collective}. Additionally, we included two popular social network datasets: Flickr~\cite{huang2017label} and Reddit~\cite{hamilton2017inductive}, along with the arxiv citation dataset from the Open Graph Benchmark (OGB)~\cite{hu2020open}. We provide more details in Appendix~\ref{data}.

{\bf Implementations.} We consider a collection of state-of-the-art GSSL methods. For contrastive methods, we include BGRL~\cite{thakoor2021large}, CCA-SSG~\cite{zhang2021canonical}, GCA~\cite{zhu2021graph}, GBT~\cite{bielak2022graph} and GraphECL~\cite{xiao2023graphecl}. For generative methods, we consider GraphMAE~\cite{hou2022graphmae}, GraphMAE2~\cite{hou2023graphmae2} and S2GAE~\cite{tan2023s2gae}. We rigorously reproduced all methods according to their papers and source codes. To ensure a fair evaluation, we perform hyperparameter tuning with the same search budget on the same dataset for all methods. More details about the implementations and the hyperparameter search process are in Appendix~\ref{model}.

\begin{table}[t]
  \caption{An overview of the datasets used in this study.}
  \label{datasets}
  \centering
  \begin{tabular}{cccccc}
    \toprule
    {\bf Dataset}     & {\bf \# Nodes}     & {\bf \#Edges}     & {\bf \# Feat.} & {\bf Avg. \# degree}& {\bf \# Classes}\\
    \midrule
    Cora & 2,708  & 5,278  & 1,433 & 3.9 & 7\\
    Citeseer     & 3,327 & 4,552  & 3,703& 2.7 & 6\\
    Pubmed     & 19,717       & 44,324  & 500& 4.5 & 3\\
    \midrule
    Flickr     & 89,250       & 899,756  & 500&10.09 &7\\
    Reddit     & 232,965       & 11,606,919  & 602&493.56 & 41\\
    ogbn-arxiv     & 169,343       & 1,166,243  & 128& 13.7 & 40\\
    \bottomrule
  \end{tabular}
\end{table}

\subsection{Research Questions}
We carefully design the \texttt{GraphFM} to systematically evaluate existing methods to motivate future research. Specifically, our aim is to address the following research questions.

{\bf RQ1: How do existing GSSL methods perform in terms of node classification performance?}

{\bf Motivation:} Node classification stands as the most commonly used task in GSSL literature. Our first research question aims to reassess existing papers within this standard task, employing consistent evaluation methods to ensure a fair comparison.

%To assess the reliability of \texttt{GraphFM}, it is necessary to train the Benchmark under standard settings, validate the reliability of the Benchmark through results, and determine which model performs better.

{\bf Experiment Design:} We conduct experiments following standard settings, wherein the models are trained on the Cora, Citeseer, and Pubmed datasets using a full-batch training strategy. Early stopping is based on accuracy for the node classification task, and performance is evaluated using the same criterion. More details can be found in Appendix~\ref{rq1}.

{\bf RQ2: How do pre-trained Graph FMs perform in terms of performance on other downstream tasks such as link prediction and node clustering?}

{\bf Motivation:} To evaluate the homogenization of GSSL methods, experiments across various downstream tasks are necessary to understand each method's generalization performance.

{\bf Experiment Design:} After obtaining pre-trained Graph FMs, we utilize the node representations post-training to conduct node classification, link prediction, and node clustering tasks. For link prediction tasks, we employ area under the curve (AUC) and average precision score (AP), while for node clustering tasks, we use normalized mutual information (NMI) and adjusted rand index (ARI), which are all the standard metrics. More details can be found in Appendix~\ref{rq2}. 
% \daochen{What is AP, is AP the same as accuracy?} \yuhao{AP is average precision score.}

{\bf RQ3: How do various training strategies (i.e., full batch, node sampling, or subgraph sampling) influence the performance of Graph FMs? How efficient are these strategies, particularly when dealing with large-scale graphs?}

{\bf Motivation:} RQ1 and RQ2 focus on small datasets, while for large-scale datasets, full-batch training strategies may not be feasible.
% , necessitating the use of mini-batch methods. 
Hence, examining model performance and efficiency under mini-batch training strategies is essential to assess scalability. 
% testing model performance and efficiency under mini-batch training strategies is crucial to verify the scalability.

{\bf Experiment Design:} We train the GSSL models on the Flickr, Reddit, and Arxiv datasets using two mini-batch training strategies: node sampling and subgraph sampling. Tasks include node classification, link prediction, and node clustering tasks. Additionally, to understand the training speed and memory usage of the GSSL methods using different sampling strategies, we report throughput and actual memory usage during training. More details can be found in Appendix~\ref{rq3}.

{\bf RQ4: Will using performances from different downstream tasks or alternative metrics as early stopping criteria impact the effectiveness of Graph FMs?}

{\bf Motivation:} In the aforementioned RQs, we save the best-performing model in node classification tasks and subsequently test it on the test set. However, the model obtained in this way may not necessarily perform well in other downstream tasks. Thus, it is essential to investigate the impact of different early stopping criteria.

{\bf Experiment Design:} We explore the viability of saving pre-trained models based on their results across different downstream tasks, such as link prediction and node clustering, and subsequently evaluate their effectiveness across various training strategies and downstream tasks. More details can be found in Appendix~\ref{rq4}. 

% \daochen{Why only save the pre-trained model through link prediction tasks, how about clustering? Should we also mention the mtrics?} \yuhao{Previously, I thought that the tasks of node classification and link prediction required training an MLP, while node clustering only required computing some mathematical functions. This led me to believe that the node representation trained by the encoder might not significantly influence the performance of node clustering. I am currently running this experiment, and if necessary, we can obtain the results very soon.}

\section{Experiments Results and Analyses}
\subsection{Performance Comparison in Node Classification (RQ1)}

We report the performance of all methods on 3 small datasets with full batch training strategy in Table~\ref{ncfull}. We made several key observations from the table.

\begin{table}[t]
    \caption{The performance of node classification for full batch in Cora, Citeseer and Pubmed. Averaged results with 5 different random seeds are reported. Highlighted are the top \textcolor{teal}{first},  \textcolor{brown}{second},  and \textcolor{blue}{third} results.}
    \label{ncfull}
    \centering
    \begin{tabular}{cccccc}
        \toprule
        Paradigm & Models & cora & citeseer & pubmed\\
        \midrule
        \multirow{5}{*}{Contrastive} & BGRL & 81.27$\pm$0.95 & 71.35$\pm$0.65 & \bf{\textcolor{brown}{86.19$\pm$0.17}} \\
         & CCA-SSG & \bf{\textcolor{teal}{86.50$\pm$0.03}} & \bf{\textcolor{brown}{73.36$\pm$0.75}} & 85.14$\pm$0.05 \\
         & GBT & 81.81$\pm$0.95 & 67.24$\pm$0.94 & 78.83$\pm$0.61 \\
         & GCA & \bf{\textcolor{brown}{85.87$\pm$0.49}} & 71.88$\pm$0.42 & \bf{\textcolor{teal}{86.22$\pm$0.75}} \\
         & GraphECL & 84.26$\pm$0.06 & 70.73$\pm$0.68 & \bf{\textcolor{blue}{86.04$\pm$0.07}} \\
        \midrule
        \multirow{3}{*}{Generative} & GraphMAE & \bf{\textcolor{blue}{85.78$\pm$0.69}} & \bf{\textcolor{teal}{73.41$\pm$0.35}} & 84.28$\pm$0.13 \\
         & GraphMAE2 & 84.84$\pm$0.22 & \bf{\textcolor{blue}{72.26$\pm$0.25}} & 84.93$\pm$0.01 \\
         & S2GAE & 82.90$\pm$0.31 & 69.34$\pm$0.19 & 81.57$\pm$0.06 \\
        \bottomrule
    \end{tabular}
\end{table}

{\bf {\textbf{\circled{\scriptsize{1}}}} Thanks to the standardized settings, our reproduced results on full-batch training are generally comparable to or sometimes even higher than those in the original paper.} \texttt{GraphFM} utilizes Optuna~\cite{akiba2019optuna} to aid in hyperparameter search for achieving optima model performance. As shown in Table~\ref{ncfull}, the top-performing datasets exhibit higher accuracy results than those reported in the original literature. Notably, the node classification results on the PubMed dataset exceed the previous benchmarks by as much as 4 percentage points. This improvement is likely due to Optuna identifying more suitable hyperparameters for the model after standardizing the settings. The only exception is that, for S2GAE, the performance is worse compared to the original paper. The possible reason is that the node classification task in the original study was conducted using an SVM classifier, whereas \texttt{GraphFM} employs an MLP head to all methods so that the results are comparable.

{\bf {\textbf{\circled{\scriptsize{2}}}} The performance gap between leading contrastive and generative paradigms on node classification is marginal.} Although the learning processes of contrastive and generative-based GSSL models differ, they exhibit similar performance on Cora, CiteSeer, and PubMed as shown in Table~\ref{ncfull}.
It is noteworthy that traditional beliefs regarding generative models (e.g., GAE~\cite{kipf2016variational}) suggest that they cannot perform comparably to contrastive-based methods on node classification tasks. However, as observed in Table~\ref{ncfull}, advanced generative models (such as GraphMAE, GraphMAE2, and S2GAE) achieve highly competitive results in classification. Particularly in the Cora and Citeseer tasks, the average performance of the generative approach even surpasses that of the contrastive models.
% The traditional generative models tended to overemphasize neighbor information at the expense of structural information, which often resulted in poorer performance in node classification tasks compared to contrastive models~\cite{Veličković_Fedus_Hamilton_Li_Bengio_Hjelm_Brain_Research_Éal}. However, as evidenced by the results in Table~\ref{ncfull}, current generative models have overcome this issue and now show improved performance in node classification tasks. 
% Particularly in the Cora and Citeseer tasks, the average performance of the generative approach is even superior to that of the contrastive models.
% especially in the Cora and Citeseer tasks, the average performance of the generative model is even better than that of the contrastive model.
%\daochen{Maybe one section corresponds to one RQ above. I feel the above RQs have some overlaps. Maybe we need to better decide/scope the RQs based on the experimental results we have}

\subsection{Performance Comparison in Link Prediction and Node Clustering (RQ2)}

In this section, we investigate the homogenization capability of pre-trained graph FMs across various tasks.
% , which examines whether pre-trained models developed through different tasks affect other downstream tasks. 
Specifically, in our experiments, \texttt{GraphFM} saves the pre-trained models based on the highest accuracy achieved in the node classification task. Subsequently, it performs three downstream tasks: node classification, link prediction, and node clustering. Since the node classification task has already been discussed in RQ1, here, we analyze the pre-trained models' generalization ability on link prediction and node clustering tasks. Figures~\ref{full_auc} and {Figure~\ref{full_nmi}} (in Appendix~\ref{clusteringfull_figure}) present the results of link prediction and node clustering, respectively, 
% display the result of link prediction and the result of node clustering can be found in Appendix~\ref{clusteringfull_figure}, 
from which we made the following observations:

\begin{figure}[t]
    \centering
    \includegraphics[scale=0.37]{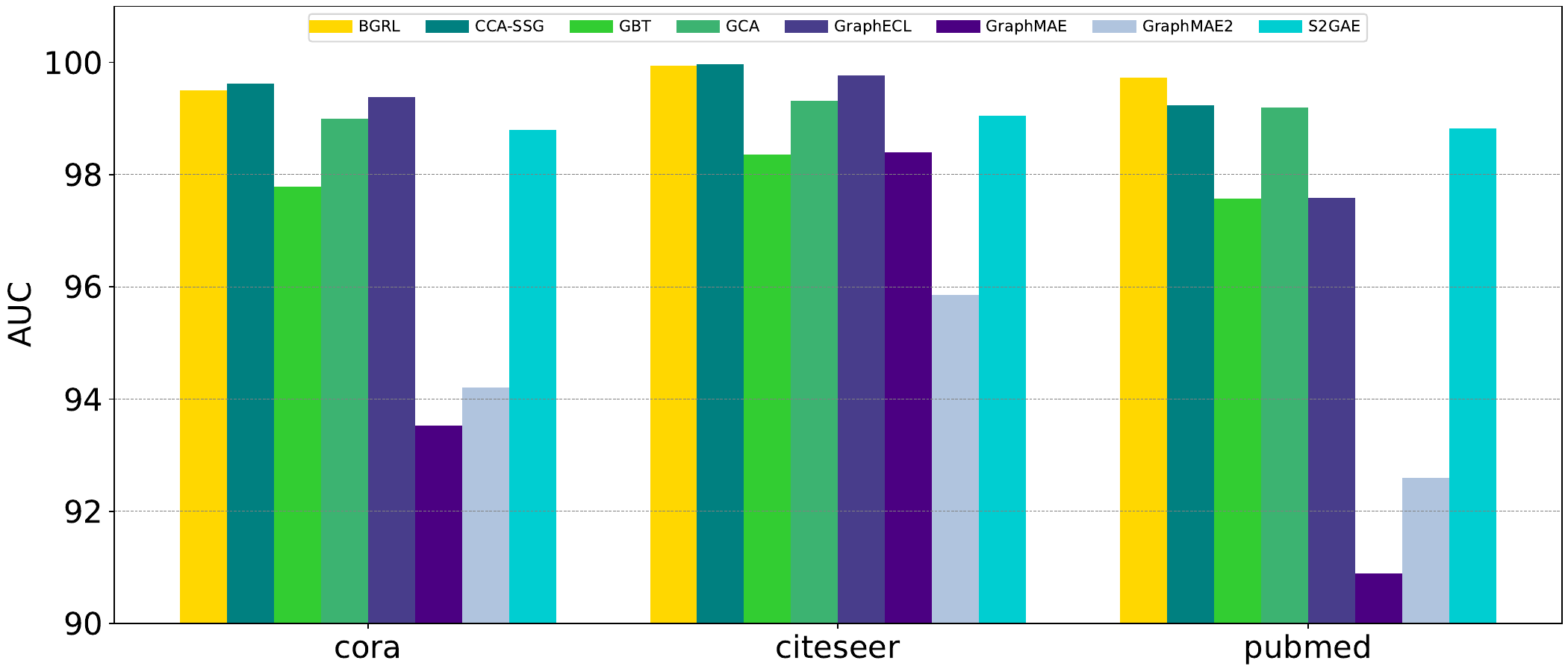}
    \caption{Link Prediction results on Cora, Citeseer, Pubmed based on full batch training. }
    \label{full_auc}
\end{figure}

{\bf {\textbf{\circled{\scriptsize{3}}}} Generative models (GraphMAE and GraphMAE2) perform poorly on link prediction tasks.} While advanced generative models like GraphMAE and GraphMAE2 demonstrate competitive performance on node classification tasks, as depicted in Table~\ref{ncfull}, they underperform other methods on link prediction tasks, as illustrated in Figure~\ref{full_auc}. The possible reason for this discrepancy is that these models solely concentrate on reconstructing node features and neglect the conventional reconstruction of network structure, which is essential for inferring missing links.

{\bf {\textbf{\circled{\scriptsize{4}}}} Although generative models fall short in link prediction, they outperform other baselines in node clustering tasks.} As depicted in Figure~\ref{full_nmi} in the Appendix~\ref{clusteringfull_figure}, except for CCA-SSG, GraphMAE and GraphMAE2 consistently surpass all contrastive-based methods on the Cora dataset and outperform all comparative methods in other scenarios. These findings, combined with their performance in node classification, underscore the advantages of node feature reconstruction as a general node-level learning objective

% which slightly outperforms the generative model on the Cora dataset, GraphMAE and GraphMAE2 excel over other models in all other instances. This indicates that these two models perform exceptionally well in reconstructing node features.

{\bf {\textbf{\circled{\scriptsize{5}}}} Both contrastive and generative models demonstrate strong homogenization capability on small datasets.} As illustrated in Figures~\ref{full_auc} and~\ref{full_nmi}, we can see that although the pre-trained models were saved according to their performance in node classification tasks, they still perform quite well across other downstream tasks in general, exhibiting good homogenization capabilities on small datasets.
% This indicates that such models exhibit good homogenization on small datasets. 
% Although the performance of node clustering on the PubMed dataset is not very well, this is primarily because PubMed only contains data for three classes and is not well-suited for this task.

\subsection{Performance and Efficiency Comparison in Large-Scale Dataset (RQ3)}

In this section, we conduct experiments across three downstream tasks to evaluate the performance of GSSL methods on large scale datasets w.r.t. different training strategies. The training results of node sampling are recorded in Table~\ref{ncnode},~\ref{lpnode},~\ref{clusteringnode}, while the training results of subgraph sampling are recorded in Table~\ref{ncsub},~\ref{lpsub},~\ref{clusteringsub}. All these tables can be found in Appendix~\ref{resultofminibatch}.

{\bf {\textbf{\circled{\scriptsize{6}}}} On small datasets, the mini-batch version of existing GSSL methods generally yields lower performance across the three downstream tasks compared to their full batch counterparts.} From Tables~\ref{ncnode} and~\ref{ncsub}, we observe that in the node classification task, the performance of almost all models decreases compared to the full batch variants, except for GBT, which shows an improvement. From Tables~\ref{lpnode} and~\ref{lpsub}, GraphMAE exhibits a significant improvement over its full batch version in the link prediction task, although contrastive models still generally perform better. From Tables~\ref{clusteringnode} and~\ref{clusteringsub}, in the node clustering task, the performance gap between contrastive and generative models diminishes compared to the full batch training scenarios. Tables~\ref{pubnode} and~\ref{pubsub} record the training results of GraphFM with mini-batch on the PubMed dataset. From the tables, we can see that, overall, the performance of mini-batch training is slightly lower compared to full-batch training. The generative models exhibit a more significant performance drop than the contrastive models.

\begin{table}[t]
    \caption{The result of \texttt{GraphFM} in Pubmed dataset with node sampling training strategy.}
    \label{pubnode}
    \centering
    \scalebox{0.87}{
    \begin{tabular}{cccc}
        \toprule
        Models & Node Classification & Link Prediction & Node Clustering \\
        \midrule
        BGRL        & 83.70$(\textcolor{green!70!black}{\uparrow2.43})$   & 99.60$(\textcolor{red}{\downarrow0.13})$/99.52$(\textcolor{red}{\downarrow0.13})$& 0.1139$(\textcolor{red}{ \downarrow0.2025 })$/0.0588$(\textcolor{red}{\downarrow0.2130})$  \\
        CCA-SSG     & 83.51$(\textcolor{red}{ \downarrow2.99 })$ & 99.58$(\textcolor{green!70!black}{ \uparrow0.35 })$/99.49$(\textcolor{green!70!black}{ \uparrow0.59 })$ & 0.1246$(\textcolor{green!70!black}{ \uparrow0.0324 })$/0.0625$(\textcolor{green!70!black}{ \uparrow0.0580 })$  \\
        GBT         & 84.23$(\textcolor{green!70!black}{ \uparrow2.42 })$   & 99.40$(\textcolor{green!70!black}{ \uparrow1.83 })$/99.31$(\textcolor{green!70!black}{ \uparrow0.41 })$ & 0.0453$(\textcolor{red}{ \downarrow0.1342 })$/-0.0084$(\textcolor{red}{ \downarrow0.0132 })$ \\
        GCA         & 82.29$(\textcolor{red}{ \downarrow3.58 })$ & 99.17$(\textcolor{red}{ \downarrow0.02 })$/99.08$(\textcolor{green!70!black}{ \uparrow0.06 })$ & 0.0708$(\textcolor{red}{ \downarrow0.1087 })$/0.0150$(\textcolor{red}{ \downarrow0.1398 })$ \\
        GraphECL    & 83.10$(\textcolor{red}{ \downarrow1.16 })$ & 95.28$(\textcolor{red}{ \downarrow2.30 })$/94.67$(\textcolor{red}{ \downarrow2.01 })$ & 0.1056$(\textcolor{green!70!black}{ \uparrow0.1676 })$/0.0199$(\textcolor{green!70!black}{ \uparrow0.2350 })$ \\
        GraphMAE    & 83.60$(\textcolor{red}{ \downarrow2.18 })$ & 96.04$(\textcolor{green!70!black}{ \uparrow5.15 })$/94.72$(\textcolor{green!70!black}{ \uparrow3.75 })$ & 0.1125$(\textcolor{red}{ \downarrow0.2107 })$/0.0250$(\textcolor{red}{ \downarrow0.2731 })$ \\
        GraphMAE2   & 80.76$(\textcolor{red}{ \downarrow4.08 })$ & 84.73$(\textcolor{red}{ \downarrow7.86 })$/85.37$(\textcolor{red}{ \downarrow6.59 })$ & 0.2770$(\textcolor{red}{ \downarrow0.0557 })$/0.2594$(\textcolor{red}{ \downarrow0.0541 })$ \\
        S2GAE       & 81.40$(\textcolor{red}{ \downarrow1.50 })$ & 88.77$(\textcolor{red}{ \downarrow10.05 })$/86.33$(\textcolor{red}{ \downarrow12.51 })$ & 0.2647$(\textcolor{green!70!black}{ \uparrow0.1671 })$/0.2757$(\textcolor{green!70!black}{ \uparrow0.2016 })$ \\
        \bottomrule
    \end{tabular}
    }
\end{table}

\begin{table}[t]
    \caption{The result of \texttt{GraphFM} in Pubmed dataset with subgraph sampling training strategy.}
    \label{pubsub}
    \centering
    \scalebox{0.87}{
    \begin{tabular}{cccc}
        \toprule
        Models & Node Classification & Link Prediction & Node Clustering \\
        \midrule
        BGRL        & 84.52$( \textcolor{red} {\downarrow1.67})$   & 99.47$(\textcolor{red} {\downarrow0.26})$/99.39$(\textcolor{red} {\downarrow0.26})$& 0.2272$(\textcolor{red} {\downarrow0.0892})$/0.1830$(\textcolor{red} {\downarrow0.0888})$  \\
        CCA-SSG     & 84.83$(\textcolor{red} {\downarrow0.31})$ & 99.69$( \textcolor{green!70!black}{\uparrow0.46})$/99.62$( \textcolor{green!70!black}{\uparrow0.72})$  & 0.2482$( \textcolor{green!70!black}{\uparrow0.1560})$/0.2214$( \textcolor{green!70!black}{\uparrow0.2169})$  \\
        GBT         & 82.61$( \textcolor{green!70!black}{\uparrow3.78})$   & 98.70$( \textcolor{green!70!black}{\uparrow1.13})$/98.83$( \textcolor{green!70!black}{\uparrow1.68})$ & 0.0638$(\textcolor{red} {\downarrow0.1157})$/-0.0018$(\textcolor{red} {\downarrow0.1566})$ \\
        GCA         & 84.43$(\textcolor{red} {\downarrow1.79})$ & 98.85$(\textcolor{red} {\downarrow0.34})$/98.80$( \textcolor{red} {\downarrow0.22})$ & 0.0888$(\textcolor{red} {\downarrow0.1844})$/0.0182$(\textcolor{red} {\downarrow0.2367})$ \\
        GraphECL    & 84.59$(\textcolor{red} {\downarrow1.45})$ & 96.23$(\textcolor{red} {\downarrow1.35})$/95.69$(\textcolor{red} {\downarrow0.99})$ & 0.3442$( \textcolor{green!70!black}{\uparrow0.2473})$/0.3057$( \textcolor{green!70!black}{\uparrow0.2878})$ \\
        GraphMAE    & 84.85$(\textcolor{green!70!black}{\uparrow0.57})$ & 95.86$( \textcolor{green!70!black}{\uparrow4.97})$/94.64$( \textcolor{green!70!black}{\uparrow3.67})$ & 0.3211$(\textcolor{red} {\downarrow0.0021})$/0.2893$(\textcolor{red} {\downarrow0.0142})$ \\
        GraphMAE2   & 80.24$(\textcolor{red} {\downarrow4.96})$ & 83.99$(\textcolor{red} {\downarrow8.60})$/84.79$(\textcolor{red} {\downarrow7.17})$ & 0.2773$(\textcolor{red} {\downarrow0.0544})$/0.2598$(\textcolor{red} {\downarrow0.0537})$ \\
        S2GAE       & 79.51$(\textcolor{red} {\downarrow2.06})$ & 89.72$(\textcolor{red} {\downarrow1.91})$/88.09$(\textcolor{red} {\downarrow3.54})$ & 0.3001$( \textcolor{green!70!black}{\uparrow0.2025})$/0.2865$( \textcolor{green!70!black}{\uparrow0.2124})$ \\
        \bottomrule
    \end{tabular}
    }
\end{table}

{\bf {\textbf{\circled{\scriptsize{7}}}} On large datasets, the performance of existing GSSL methods varies significantly across different tasks.} In the node classification task, both contrastive and generative-based methods exhibit similar performance across the three datasets. However, in the node clustering task, generative models consistently outperform contrastive models in all cases. It is noteworthy that some results display negative values in this task. This arises from the calculation formula of the Adjusted Rand Index (ARI), which ranges from [-1, 1]. Thus, these negative values fall within the expected range. Table~\ref{Flikr_node_subgraph} shows the results of \texttt{GraphFM} in Flickr dataset.

For training efficiency, {\bf {\textbf{\circled{\scriptsize{8}}}} generative models do not encounter out-of-memory issues, providing them with a notable advantage in scalability on large-scale datasets.} Comparing the two mini-batch methods, {\bf {\textbf{\circled{\scriptsize{9}}}} subgraph sampling exhibits the lowest memory usage and the fastest training speed,} as evidenced in Table~\ref{efficiency}, Table~\ref{efficiency_minibatch} (in Appendix~\ref{resultsofefficiency}), and Figure~\ref{pubmed_mem}, particularly on larger datasets. In summary, considering the variable performance of mini-batch variants, exploring the design of effective self-supervised training architectures or objectives within the mini-batch framework represents a promising avenue for future research.

\begin{table}[t]
    \caption{The result of \texttt{GraphFM} in Flickr dataset. " - " means out of memory.}
    \label{Flikr_node_subgraph}
    \centering
    \begin{tabular}{ccccc}
        \toprule
        Training Strategy& Models & Node Classification & Link Prediction & Node Clustering \\
        \midrule
        \multirow{8}{*}{Node Sampling}&BGRL & 47.37$\pm$0.05 & 87.88/88.24 & 0.0054/0.0145  \\
        &CCA-SSG & 51.59$\pm$0.14 & 76.45/14.44 & 0.0622/0.0397  \\
        &GBT & 52.11$\pm$0.08 & 86.69/87.93 & 0.0179/0.0175 \\
        &GCA & - & - &  - \\
        &GraphECL & - & - &  - \\
        \cmidrule(r){2-5}
        &GraphMAE & 49.25$\pm$0.13  & 50.00/50.00 & 0.0154/0.0197 \\
        &GraphMAE2 & 46.07$\pm$0.83 & 49.94/49.98 & 0.0157/0.0097 \\
        &S2GAE & 43.90$\pm$0.17 & 49.95/49.93 & 0.0067/0.0054 \\
        \midrule
        \multirow{8}{*}{Subgraph Sampling}& BGRL & 47.14$\pm$0.07 & 86.92/87.57 & 0.0052/0.0145 \\
        &CCA-SSG & 50.95$\pm$0.20 & 50.00/50.00 & 0.0181/0.0125  \\
        &GBT & 51.00$\pm$0.17  & 50.00/50.00 & 0.0453/0.0176  \\
        &GCA & 51.76$\pm$0.08 & 50.00/50.00 & 0.0343/0.0155  \\
        &GraphECL & 46.23$\pm$0.09 & 51.06/51.12 &  0.0128/0.0142 \\
        \cmidrule(r){2-5}
        &GraphMAE & 45.30$\pm$0.85 & 50.00/50.00 & 0.0090/0.0137  \\
        &GraphMAE2 & 46.25$\pm$0.90 & 49.94/49.98 & 0.0157/0.0068 \\
        &S2GAE & 43.97$\pm$0.20 & 49.91/49.91 & 0.0039/0.0033 \\
        \bottomrule
    \end{tabular}
\end{table}

\begin{figure}[t]
    \centering
    \includegraphics[scale=0.3]{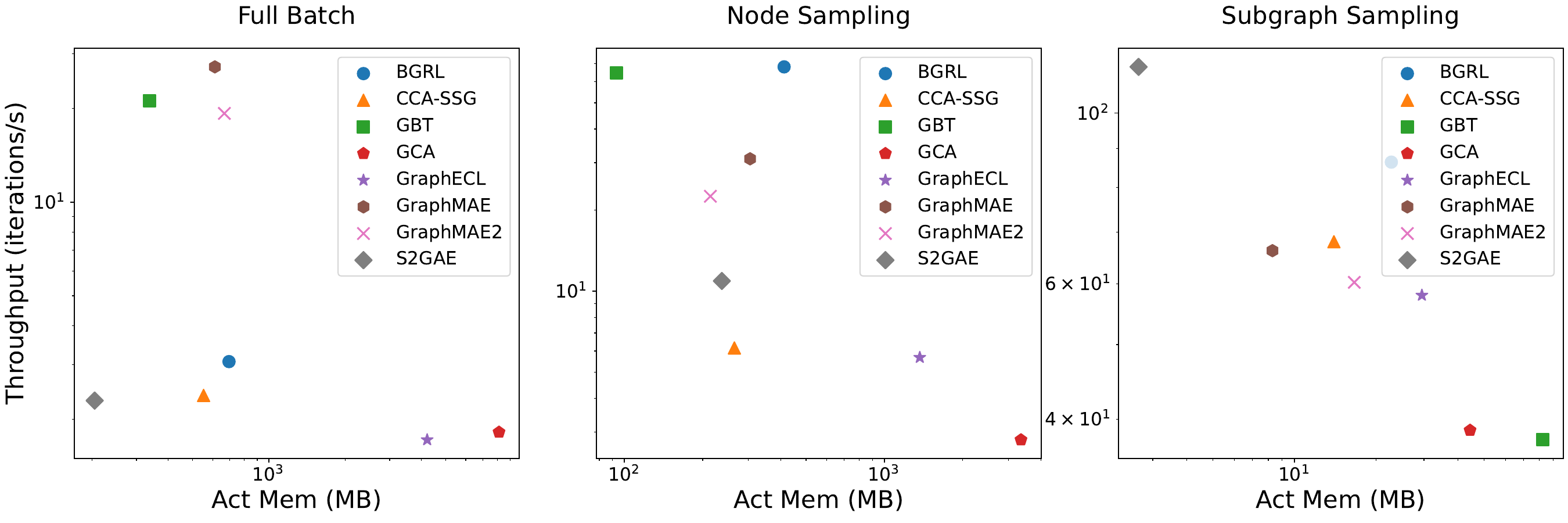}
    \caption{Time and space consumption of different methods and training strategy on Pubmed.}
    \label{pubmed_mem}
\end{figure}

\subsection{Performance Using Alternative Early Stopping Criterion (RQ4)}
Since all models perform well with both full batch and mini-batch methods on small datasets, our experiment will focus on large-scale dataset. In this section, we focus on using link prediction and node clustering as the early stop criteria. Specifically, 
% —we focus on the Flickr dataset in this experiment. The results on the Flickr dataset are evidently more anomalous compared to other datasets. Therefore, in this experiment, 
we save pre-trained models based on their performance in the downstream task and test there performance across three different downstream tasks on large dataset. Table~\ref{flp} reports the results on Flickr dataset, and more results can be found in Appendix~\ref{stopcri}.

{\bf {\textbf{\circled{\scriptsize{10}}}} GSSL methods can achieve better performance on a downstream task when using the same task as the early stop criteria.} According to Table~\ref{flp}, it is evident that compared to previous experiments, the performance of contrastive models in link prediction has significantly improved. Conversely, no substantial performance enhancement is observed in generative methods. These findings underscore the impact of early stopping criteria on various self-supervised training methodologies, especially for contrastive-based approach. Similar observations can be made when employing node clustering as the early stoping criterion (see Appendix~\ref{stopcri} for details).
%And the experimental results for node clustering are similar to those for node classification.

% by using link prediction result as the early stop indicator. We observed that when using link prediction as the criterion for saving pre-trained models, the performance in node classification and node clustering tasks shows little difference compared to previous results. This may be due to inherent issues with the Flickr dataset, making it difficult to capture node features effectively. However, most contrastive models show significant improvement in link prediction performance. On the other hand, generative models still perform poorly in link prediction.

\begin{table}[t]
    \caption{The result of \texttt{GraphFM} in Flickr dataset by saving valid model with the best performance in link prediction. " - " means out of memory.}
    \label{flp}
    \centering

    \begin{tabular}{ccccc}
        \toprule
        Training Strategy& Models & Node Classification & Link Prediction & Node Clustering \\
        \midrule
        \multirow{8}{*}{Node Sampling}&BGRL & 46.07$\pm$0.56 & 86.60/87.31 & 0.0073/0.0195  \\
        &CCA-SSG & 51.03$\pm$0.03 & 98.98/98.90 & 0.0555/0.0538  \\
        &GBT & 51.66$\pm$0.17 & 85.39/86.64 & 0.0727/0.0386 \\
        &GCA & - & - &  - \\
        &GraphECL & - & - &  - \\
        \cmidrule(r){2-5}
        &GraphMAE & 45.86$\pm$0.05 & 50.00/50.00 & 0.0128/0.0082 \\
        &GraphMAE2 & 45.80$\pm$0.18 & 50.08/50.04 & 0.0257/0.0090 \\
        &S2GAE & 45.13$\pm$0.32 & 50.65/50.41 & 0.0246/0.0175 \\
        \midrule
        \multirow{8}{*}{Subgraph Sampling}&BGRL & 46.09$\pm$1.31 & 86.36/87.06 &  0.0074/0.0211 \\
        &CCA-SSG & 50.92$\pm$0.05 & 86.36/87.50 & 0.0827/0.0645  \\
        &GBT & 52.14$\pm$0.06 & 52.24/57.72 & 0.0423/0.0134 \\
        &GCA & 51.94$\pm$0.72 & 72.84/65.80 & 0.0744/0.0315  \\
        &GraphECL & 47.09$\pm$0.34 & 51.16/50.92 &  0.0043/0.0022 \\
        \cmidrule(r){2-5}
        &GraphMAE & 49.63$\pm$0.42 & 57.67/54.80 & 0.0201/0.0124 \\
        &GraphMAE2 & 45.86$\pm$0.09 & 50.10/50.05 & 0.0249/0.0109 \\
        &S2GAE & 44.76$\pm$0.43 & 50.25/50.30 & 0.0068/0.0049 \\
        \bottomrule
    \end{tabular}
\end{table}

\section{Future Direrctions}

Drawing upon our empirical analyses, we point out some promising future directions for \texttt{GraphFM}.

{\bf Reconsidering the homogeneity of contrastive models and generative models is imperative.} Homogeneity is a significant characteristic of FMs and should be given high priority. However, based on the results from {\texttt{GraphFM}, current contrastive and generative models face substantial challenges in achieving homogeneity. These challenges arise from various factors such as the datasets, node-level or edge-level downstream tasks.

{\bf Exploring an effective early stop strategy for GNN pre-training.} Based on the above experiments, no single early stopping criterion currently enhances model performance across various downstream tasks, contradicting the original intention of the foundation model. Future research should focus on exploring more effective early stopping criteria.

% {\bf Cross-domain graph foundation model trianing.} The experiments detailed in this study predominantly focus on training with a single dataset. However, training with cross-domain datasets has the potential to enhance model consistency. It is advisable not only to train the model using datasets within the same domain but also to evaluate its performance using datasets from diverse domains.

{\bf How to extend graph foundation model to textual attributed graphs.} Presently, our training primarily revolves around conventional graph datasets, where where node features are numerical vectors. Nonetheless, in real-world graph applications, nodes are often characterized by textual descriptions, such as social media posts on Twitter, formally referred to as textual attribute graphs. Whether GraphFM can be extended to accommodate such graphs, and how it should be extended, remains an open research question.

% {\bf Can large language models (LLMs) be used to assist in training.}

\section{Conclusion}
This paper introduces \texttt{GraphFM}, a comprehensive benchmark for Graph Foundation Models. We reimplement and compare 8 leading GSSL methods across diverse datasets, providing a fair comparison and insightful analysis into this burgeoning research field. Our empirical observations reveal variations in performance between full-batch and mini-batch training scenarios. Furthermore, we find that existing self-supervised GNN pre-training efforts may not effectively serve as foundation models on graphs, as they often struggle to generalize well across key graph reasoning tasks (node classification, link prediction, and node clustering) simultaneously. Notably, we highlight the significant impact of early stopping criteria in GNN pre-training on model generalization capability, a critical issue previously overlooked by the research community. We believe that this benchmark will have a positive impact on this emerging research domain. Our code is publicly available, and we encourage contributions of new datasets and methods. 
In the future, we aim to extend the applicability of \texttt{GraphGLM} to text-attributed graphs and broaden its support for various graph-level learning tasks and heterogeneous graphs, enhancing its versatility and comprehensiveness.

% \section*{Acknowledge}

%%%%%%%%%%%%%%%%%%%%%%%%%%%%%%%%%%%%%%%%%%%%%%%%%%%%%%%%%%%%

\bibliographystyle{unsrt}
\bibliography{ref}

%%%%%%%%%%%%%%%%%%%%%%%%%%%%%%%%%%%%%%%%%%%%%%%%%%%%%%%%%%%%

%%%%%%%%%%%%%%%%%%%%%%%%%%%%%%%%%%%%%%%%%%%%%%%%%%%%%%%%%%%%

%%%%%%%%%%%%%%%%%%%%%%%%%%%%%%%%%%%%%%%%%%%%%%%%%%%%%%%%%%%%

\newpage

\appendix

\section{Additional Details on Benchmark}

\subsection{Datasets}
\label{data}
{\bf Cora, Citeseer and Pubmed}~\cite{sen2008collective} are three citation networks commonly used in prior GSSL works~\cite{thakoor2021large,zhang2021canonical,xiao2023graphecl,hou2022graphmae,hou2023graphmae2,tan2023s2gae}. In these datasets, nodes represent academic papers, and edges denote citation relationships between the papers. Each node's features are represented as bag-of-words vectors, and the label assigned to each node corresponds to its research topic category.

{\bf Flickr}~\cite{huang2017label} is a social network dataset where nodes represent users and edges represent interactions between users (such as comments and likes). Node features are metadata attributes derived from users' photos. The label of each node is not predefined, making it suitable for tasks like link prediction and node clustering.

{\bf Reddit}~\cite{hamilton2017inductive} is a social network dataset where nodes represent posts and edges represent comments linking the posts. Node features are 602-dimensional vectors representing various attributes of the posts, such as word embeddings. The label of each node corresponds to the community or subreddit to which the post belongs, with 41 different classes in total.

{\bf Ogbn-arxiv}~\cite{hu2020open} is a citation network dataset from the Open Graph Benchmark (OGB) suite. Nodes represent papers from the arXiv repository, and edges represent citation relationships between papers. Node features are 128-dimensional vectors representing word2vec embeddings of paper abstracts. The label of each node is the subject area of the paper, with 40 different categories in total.

\subsection{GSSL models}
\label{model}

{\bf BGRL}~\cite{thakoor2021large} is a contrastive learning model that focuses on learning node representations by maximizing agreement between different views of the same graph. It leverages bootstrapping techniques to create positive and negative samples, ensuring robust and informative embeddings.

{\bf CCA-SSG}~\cite{zhang2021canonical} applies canonical correlation analysis to graph data for self-supervised learning. The method aims to find representations that maximize the correlation between two sets of views from the graph, promoting the extraction of common features and enhancing the quality of node embeddings.

{\bf GBT}~\cite{bielak2022graph} is a self-supervised learning model specifically designed for graph-structured data. Inspired by the Barlow Twins framework from computer vision, GBT aims to learn meaningful node representations by maximizing the similarity between different augmented views of the same graph while minimizing redundancy between feature dimensions.

{\bf GCA}~\cite{zhu2021graph} is a graph contrastive learning method that generates augmented views of the graph and uses these views to learn node representations. It optimizes the agreement between the embeddings of the original and augmented graphs, helping the model to generalize better across different tasks.

{\bf GraphECL}~\cite{xiao2023graphecl} is an advanced contrastive learning model that enhances the basic framework by incorporating additional graph structural information. It improves the quality of learned embeddings by leveraging both node attributes and structural features, making it effective for various graph-based tasks.

{\bf GraphMAE}~\cite{hou2022graphmae} is inspired by the success of masked autoencoders in NLP. It masks a portion of the graph data (such as node features or edges) and trains the model to reconstruct the masked parts. This approach helps in learning robust and informative node representations without relying on labeled data.

{\bf GraphMAE2}~\cite{hou2023graphmae2} builds on the original GraphMAE, introducing enhancements to the masking and reconstruction mechanisms. It may involve more sophisticated masking strategies, improved network architectures, or additional training objectives to further enhance the quality of learned embeddings.

{\bf S2GAE}~\cite{tan2023s2gae} is a generative model that uses autoencoders for graph data. It employs self-supervised learning techniques to train the autoencoder to reconstruct the graph from its latent representation. This process helps in capturing the underlying structure and features of the graph, making the embeddings useful for downstream tasks like node classification and clustering.

\begin{table}[h!]
    \caption{Hyper-parameter search space of all implemented methods.}
    \label{Hypsearch}
    \centering
    \begin{tabular}{c|l|l}
        \toprule
        Models  & Hyper-parameter & Search Space \\
        \midrule
        \midrule
        General Settings & lr & [1e-6, 1e-2] \\
         & weight\_decay & [1e-6, 1e-2] \\
         & batch\_size & 512, 1024, 2048, 4096, 10000, 20000 \\
         & decode\_channels\_lp & 128, 256, 512, 1024 \\
         & decode\_layers\_lp & 1, 2, 4, 8 \\
         \midrule
        BGRL~\cite{thakoor2021large} & drop\_edge\_p\_1 & 0.0, 0.1, 0.2, 0.3, 0.4, 0.5, 0.6 \\
         & drop\_edge\_p\_2 & 0.0, 0.1, 0.2, 0.3, 0.4, 0.5, 0.6 \\
         & drop\_feat\_p\_1 & 0.0, 0.1, 0.2, 0.3, 0.4, 0.5, 0.6 \\
         & drop\_feat\_p\_2 & 0.0, 0.1, 0.2, 0.3, 0.4, 0.5, 0.6 \\
         \midrule
        CCA-SSG~\cite{zhang2021canonical} & dfr & 0.0, 0.1, 0.2, 0.3, 0.4, 0.5, 0.6 \\
         & der & 0.0, 0.1, 0.2, 0.3, 0.4, 0.5, 0.6 \\
         & hid\_dim & 128, 256, 512, 1024 \\
         \midrule
        GBT~\cite{zhu2021graph} & emb\_dim & 128, 256, 512, 1024 \\
         & lr\_base & [1e-6, 1e-2] \\
         & p\_x & 0.0, 0.1, 0.2, 0.3, 0.4, 0.5, 0.6 \\
         & p\_e & 0.0, 0.1, 0.2, 0.3, 0.4, 0.5, 0.6 \\
         \midrule
        GCA~\cite{bielak2022graph} & num\_hidden & 128, 256, 512, 1024 \\
         & drop\_edge\_rate\_1 & 0.0, 0.1, 0.2, 0.3, 0.4, 0.5, 0.6 \\
         & drop\_edge\_rate\_2 & 0.0, 0.1, 0.2, 0.3, 0.4, 0.5, 0.6 \\
         & drop\_feature\_rate\_1 & 0.0, 0.1, 0.2, 0.3, 0.4, 0.5, 0.6 \\
         & drop\_feature\_rate\_2 & 0.0, 0.1, 0.2, 0.3, 0.4, 0.5, 0.6 \\
         \midrule
        GraphECL~\cite{xiao2023graphecl} & hid\_dim & 128, 256, 512, 1024, 2048 \\
         & n\_layers & [1, 4] \\
         & temp & 0.4, 0.5, 0.6, 0.7, 0.8 \\
         & lam & [1e-6, 1e-2] \\
        \midrule
        GraphMAE~\cite{hou2022graphmae}
         & num\_heads & 1, 2, 4, 8 \\
         & num\_hidden & 256, 512, 1024 \\
         & attn\_drop & 0.0, 0.1, 0.2, 0.3, 0.4, 0.5 \\
         & in\_drop & 0.0, 0.1, 0.2, 0.3, 0.4, 0.5 \\
         & negative\_slope & 0.0, 0.1, 0.2, 0.3, 0.4, 0.5 \\
         & mask\_rate & 0.4, 0.5, 0.6, 0.7, 0.8 \\
         & drop\_edge\_rate & 0.0, 0.05, 0.15, 0.20 \\
         & $alpha_l$ & 1, 2, 3 \\
         \midrule
        GraphMAE2~\cite{hou2023graphmae2} 
         & num\_heads & 1, 2, 4, 8 \\
         & num\_hidden & 256, 512, 1024 \\
         & attn\_drop & 0.0, 0.1, 0.2, 0.3, 0.4, 0.5 \\
         & in\_drop & 0.0, 0.1, 0.2, 0.3, 0.4, 0.5 \\
         & negative\_slope & 0.0, 0.1, 0.2, 0.3, 0.4, 0.5 \\
         & mask\_rate & 0.4, 0.5, 0.6, 0.7, 0.8 \\
         & drop\_edge\_rate & 0.0, 0.05, 0.15, 0.20 \\
         & $alpha_l$ & 1, 2, 3 \\
         & replace\_rate & 0.0, 0.1, 0.2, 0.3, 0.4, 0.5 \\
         & lam & 0.0, 0.1, 0.2, 0.3, 0.4, 0.5 \\
         \midrule
        S2GAE~\cite{tan2023s2gae} & dim\_hidden & 128, 256, 512, 1024 \\
         & decode\_channels & 128, 256, 512, 1024 \\
         & decode\_layers & [1, 8] \\
         & mask\_ratio & 0.4, 0.5, 0.6, 0.7, 0.8 \\
        \bottomrule
    \end{tabular}
\end{table}

\section{Additional Experimental Details}

\subsection{RQ1}
\label{rq1}
{\bf General Experimental Settings.} We strive to adhere to the original implementation of various GSSL models provided in their provided in their respective papers or source codes. To achieve this, we have integrated different options into a standardized framework as shown in Figure~\ref{architecture}. To ensure fairness and consistency, we have standardized the optimizer as well as the evaluation methods for node classification, link prediction, and node clustering. Additionally, we have adopted the method of splitting edges for link prediction and adhered to the data splitting approach used in PyG~\cite{fey2019fast}.

{\bf Hyperparameter.} We conduct comprehensive hyperparameter tuning through Optuna~\cite{akiba2019optuna} to ensure a thorough and impartial evaluation of these GSSL models. The hyperparameter search spaces of all models are presented in Table~\ref{Hypsearch}, the notation "[]" indicates the range for hyperparameter tuning, while the absence of brackets denotes specific values used in the search. For detailed meanings of these hyperparameters, please refer to their original papers.

\subsection{RQ2}
\label{rq2}
In our link prediction tasks, we use AUC (Area Under the Curve) and AP (Average Precision) as metrics, while for node clustering, we employ NMI (Normalized Mutual Information) and ARI (Adjusted Rand Index)~\cite{vinh2009information}. These metrics are widely recognized as effective for these respective tasks~\cite{shi2023gigamae}. The following are the details for these metrics.

\subsubsection{AUC}
AUC measures the ability of the model to distinguish between positive and negative edges. It is calculated as the area under the Receiver Operating Characteristic (ROC) curve.

\begin{equation}
\text{AUC} = \int_{0}^{1} \text{TPR}(FPR) \, d(\text{FPR}),
\end{equation}

where \text{TPR} (True Positive Rate) and \text{FPR} (False Positive Rate) are defined as:
\[
\text{TPR} = \frac{\text{TP}}{\text{TP} + \text{FN}}
\]
\[
\text{FPR} = \frac{\text{FP}}{\text{FP} + \text{TN}}
\]

\subsubsection{AP}
AP summarizes a precision-recall curve as the weighted mean of precisions achieved at each threshold, with the increase in recall from the previous threshold used as the weight.

\begin{equation}
\text{AP} = \sum_{n} (R_n - R_{n-1}) P_n,
\end{equation}

where \( P_n \) and \( R_n \) are the precision and recall at the \( n \)-th threshold.

\subsubsection{NMI}
NMI measures the similarity between the clustering of the nodes and the ground truth labels. It is defined as:

\begin{equation}
\text{NMI}(U, V) = \frac{I(U; V)}{\sqrt{H(U) H(V)}},
\end{equation}

where \( I(U; V) \) is the mutual information between the cluster assignments \( U \) and \( V \), and \( H(U) \) and \( H(V) \) are the entropies of \( U \) and \( V \), respectively.

\[
I(U; V) = \sum_{u \in U} \sum_{v \in V} p(u, v) \log \frac{p(u, v)}{p(u) p(v)}
\]
\[
H(U) = - \sum_{u \in U} p(u) \log p(u)
\]

\subsubsection{ARI}
The ARI measures the similarity between two data clusterings, corrected for chance. It is defined as:

\begin{equation}
\text{ARI} = \frac{\sum_{ij} \binom{n_{ij}}{2} - \left[ \sum_i \binom{a_i}{2} \sum_j \binom{b_j}{2} \right] / \binom{n}{2}}{0.5 \left[ \sum_i \binom{a_i}{2} + \sum_j \binom{b_j}{2} \right] - \left[ \sum_i \binom{a_i}{2} \sum_j \binom{b_j}{2} \right] / \binom{n}{2}},
\end{equation}

where \( n_{ij} \) is the number of elements in both cluster \( i \) of the true clustering and cluster \( j \) of the predicted clustering, \( a_i \) is the number of elements in cluster \( i \), \( b_j \) is the number of elements in cluster \( j \), and \( n \) is the total number of elements.

\subsection{RQ3}
\label{rq3}
In our experiments, we selected Node Sampling~\cite{hamilton2017inductive} and Subgraph Sampling~\cite{chiang2019cluster} two sampling strategies. The formula mentioned in Section~\ref{sec:2} is the general formula for message passing. In the mini-batch setting, the function will be as follows:
\[
\mathbf{X}_{\mathcal{B}_0}^{(k)} = \tilde{\mathbf{A}}_{\mathcal{B}_1}^{(k-1)} \sigma \left( \tilde{\mathbf{A}}_{\mathcal{B}_2}^{(k-2)} \sigma \left( \cdots \sigma \left(\tilde{\mathbf{A}}_{\mathcal{B}_K}^{(0)} \mathbf{X}_{\mathcal{B}_K}^{(0)} \mathbf{W}^{(0)} \right) \cdots \right) \mathbf{W}^{(K-2)} \right) \mathbf{W}^{(K-1)}
\]
where $\mathcal{B}_l$ is the set of sampled nodes for the $l$-th layer, and $\tilde{\mathbf{A}}^{(l)}$ is the adjacency matrix for the $l$-th layer sampled from the full graph. The key difference among different sampling methods is how $\{\mathcal{B}_0, \ldots, \mathcal{B}_{K-1}, \mathcal{B}_K\}$ are sampled, the following are the details for these two methods.

\subsubsection{Node Sampling}
\(\mathcal{B}_{l+1} = \bigcup_{v \in \mathcal{B}_l} \{u \mid u \sim Q \cdot \mathbb{P}_{\mathcal{N}(v)}\}\), where \(\mathbb{P}\) is a uniform distribution; \(\mathcal{N}(v)\) is the sampling space, i.e., the 1$-$hop neighbors of \(v\); and \(Q\) denotes the number of samples.

\subsubsection{Subgraph Sampling}
\(\mathcal{B}_K = \mathcal{B}_{K-1} = \cdots = \mathcal{B}_0 = \{u \mid u \sim Q \cdot \mathbb{P}_{\mathcal{G}}\}\). In the subgraph-wise sampling, all layers share the same subgraph induced from the entire graph \(G\) based on a specific sampling strategy \(\mathbb{P}_{\mathcal{G}}\), such that the sampled nodes are confined in the subgraph. ClusterGCN~\cite{chiang2019cluster} first partitions the entire graph into clusters based on some graph partition algorithms, e.g., METIS~\cite{karypis1998fast}, and then selects several clusters to form a batch.

\subsection{RQ4}
\label{rq4}
To explore the impact of different metrics as early stopping criteria on model performance, we conducted experiments by replacing the usual accuracy metric in node classification with AUC for link prediction and NMI for node clustering.

\section{Additional Results}
\label{results}
{\bf Running Experiments.} Our experiments are mostly conducted on a Linux server with Lenovo SR670, and an NVIDIA RTX8000 GPU (48G).

\subsection{Result of Full Batch}
\label{clusteringfull_figure}

In Section 4.2, we analyzed the outcomes of various downstream tasks that differ from saving the best validation model. The results of node clustering are recorded in Figure~\ref{full_nmi}, and the analysis is detailed in Section 4.2.
% \begin{table}[t]
%     \caption{Link prediction, Full batch (AUC/AP)}
%     \label{lpfull}
%     \centering
%     \begin{tabular}{cccc}
%         \toprule
%         Models  & Cora & Citeseer & Pubmed \\
%         \midrule
%         BGRL & {\bf \textcolor{brown}{99.50}}/{\bf \textcolor{brown}{99.53}} & 99.94/99.94 & 99.73/99.65\\
%         CCA-SSG &  {\bf \textcolor{teal}{99.62}}/{\bf \textcolor{teal}{99.56}} & {\bf \textcolor{teal}{99.97}}/{\bf \textcolor{teal}{99.96}} & 99.23/98.90 \\
%         GBT & 97.79/97.33 & 98.36/98.13 & 97.57/97.15 \\
%         GCA & 98.99/98.76 & 99.31/99.30 & 99.19/99.02 \\
%         GraphECL & {\bf \textcolor{blue}{99.38}}/98.79 & 99.77/99.73 & 97.58/96.68 \\
%         \midrule
%         GraphMAE & 93.53/93.07 & 98.39/98.45 & 90.89/90.97 \\
%         GraphMAE2 & 94.21/94.73 & 95.85/96.48 & 92.59/91.96 \\
%         S2GAE & 98.80/{\bf \textcolor{blue}{98.91}} & 99.05/99.14 & 98.82/98.44 \\
%         \bottomrule
%     \end{tabular}
% \end{table}

\begin{figure}[h!]
    \centering
    \includegraphics[scale=0.37]{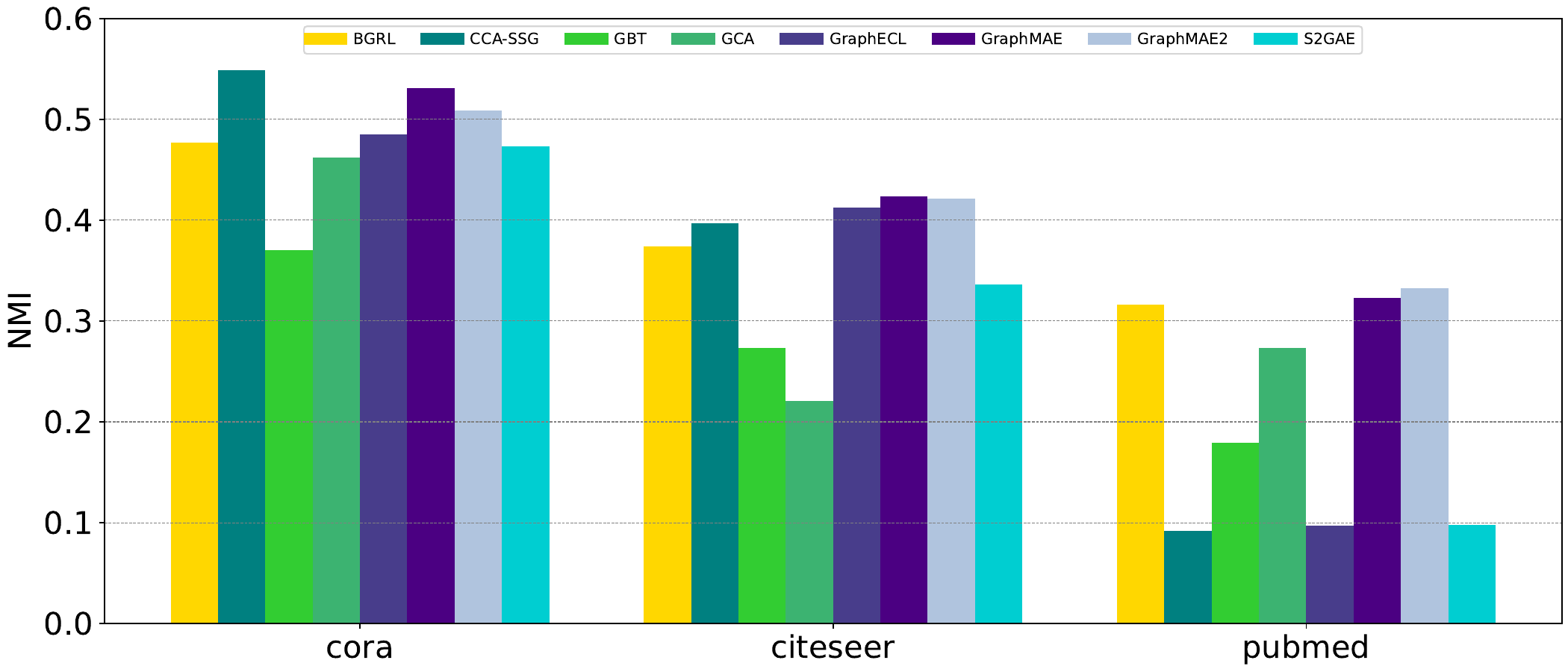}
    \caption{Node Clustering results on Cora, Citeseer, Pubmed based on full batch training. }
    \label{full_nmi}
\end{figure}

% \begin{table}[t]
%     \caption{Node clustering, Full batch (NMI/ARI)}
%     \label{clusteringfull}
%     \centering
%     \begin{tabular}{cccc}
%         \toprule
%         Models & Cora & Citeseer & Pubmed\\
%         \midrule
%         BGRL & 0.4769/0.3612 & 0.3743/0.3349 & 0.3164/0.2718 \\
%         CCA-SSG & 0.5488/0.5051 & 0.3968/0.3323 & 0.0922/0.0045 \\
%         GBT & 0.3703/0.2630 & 0.2736/0.2671 & 0.1795/0.1548 \\
%         GCA & 0.4625/0.3849 & 0.2209/0.1948 & 0.2732/0.2549 \\
%         GraphECL & 0.4853/0.3622 & 0.4128/0.4105 & 0.0969/0.0179 \\
%         \midrule
%         GraphMAE & 0.5309/0.4278 & 0.4236/0.4228 & 0.3232/0.2981 \\
%         GraphMAE2 & 0.5089/0.3778 & 0.4215/0.4183 & 0.3327/0.3135 \\
%         S2GAE & 0.4729/0.3571 & 0.3365/0.2893 & 0.0976/0.0741\\
%         \bottomrule
%     \end{tabular}
% \end{table}

\subsection{Result of Mini-Batch}
\label{resultofminibatch}
In Section 4.3, we analyzed the performance of \texttt{GraphFM} using the mini-batch training strategy across various datasets. Additional results are recorded in Tables~\ref{ncnode},~\ref{ncsub},~\ref{lpnode},~\ref{lpsub},~\ref{clusteringnode} and~\ref{clusteringsub}. 

From the table, we can observe that S2GAE, through its mini-batch training strategy, does not perform as well as other models in node classification on large-scale datasets. However, it typically performs better in link prediction, and its performance in node clustering tasks is not significantly different from other models. This indicates that S2GAE's model scalability is related to the downstream tasks, exhibiting stronger scalability in link prediction tasks.

\begin{table}[h!]
    \caption{The result of \texttt{GraphFM} in Node classification with Node sampling. " - " means out of memory.}
    \label{ncnode}
    \centering
    \scalebox{0.87}{
    \begin{tabular}{cccccccc}
        \toprule
        Models & cora & citeseer & pubmed & Flickr & Reddit & Arxiv\\
        \midrule
        BGRL & 83.69$\pm$0.20 & 70.12$\pm$0.53 & 83.70$\pm$0.07 & 47.37$\pm$0.05 & 93.16$\pm$0.03 & 65.20$\pm$0.57 \\
        CCA-SSG & 86.29$\pm$0.18 & 71.58$\pm$0.13 & 83.51$\pm$0.14 & 51.59$\pm$0.14 & 93.51$\pm$0.86 & 67.16$\pm$0.18 \\
        GBT & 82.71$\pm$0.64 & 68.56$\pm$1.32 & 84.23$\pm$0.19 & 52.11$\pm$0.08 & 92.17$\pm$0.09 & 61.60$\pm$0.25\\
        GCA & 86.05$\pm$0.20 & 72.92$\pm$0.52 & 82.29$\pm$0.14 & - & - & -\\
        GraphECL & 78.11$\pm$0.25 & 65.52$\pm$0.23 & 83.10$\pm$0.04 & - & - & -\\
        \midrule
        GraphMAE & 83.96$\pm$0.69 & 70.63$\pm$0.18 & 83.60$\pm$0.06 & 49.25$\pm$0.13 & 94.30$\pm$0.04 & 66.33$\pm$0.11 \\
        GraphMAE2 & 77.52$\pm$0.52 & 64.77$\pm$0.78 & 80.76$\pm$0.18 & 46.07$\pm$0.83 & 92.66$\pm$0.30 & 65.45$\pm$0.01 \\
        S2GAE & 77.22$\pm$0.88 & 64.85$\pm$1.25 & 81.40$\pm$0.37 & 43.90$\pm$0.17 & 62.73$\pm$0.84 & 56.58$\pm$0.71 \\
        \bottomrule
    \end{tabular}}
\end{table}

\begin{table}[h!]
    \caption{The result of \texttt{GraphFM} in Node classification with Subgraph sampling. " - " means out of memory.}
    \label{ncsub}
    \centering
    \scalebox{0.87}{
    \begin{tabular}{cccccccc}
        \toprule
        Models & Cora & Citeseer & Pubmed & Flickr & Reddit & Arxiv \\
        \midrule
        BGRL & 83.69$\pm$0.70 & 70.75$\pm$0.41 & 84.52$\pm$0.14 & 47.14$\pm$0.07 & 93.39$\pm$0.08 & 64.68$\pm$0.12 \\
        CCA-SSG & 84.90$\pm$0.95 & 71.96$\pm$0.25 & 84.83$\pm$0.16 & 50.95$\pm$0.20 & 93.53$\pm$0.42 & 66.62$\pm$0.17\\
        GBT & 83.09$\pm$0.61 & 68.11$\pm$1.59 & 82.61$\pm$0.13 & 51.00$\pm$0.17 & 92.25$\pm$0.11 & 60.10$\pm$0.04\\
        GCA & 85.71$\pm$0.26 & 70.24$\pm$0.54 & 84.43$\pm$0.31 & 51.76$\pm$0.08 & 91.46$\pm$0.33 & 62.87$\pm$0.43\\
        GraphECL & 84.29$\pm$0.46 & 72.16$\pm$0.20 & 84.59$\pm$0.37 & 46.23$\pm$0.09 & - & 58.51$\pm$0.11\\
        \midrule
        GraphMAE & 85.72$\pm$0.77 & 72.51$\pm$0.50 & 84.85$\pm$0.17 & 45.30$\pm$0.85 & 94.41$\pm$0.14 & 67.33$\pm$0.05 \\
        GraphMAE2 & 78.32$\pm$1.01 & 64.45$\pm$0.35 & 80.24$\pm$0.25 & 46.25$\pm$0.90 & 91.86$\pm$0.58 & 65.71$\pm$0.01 \\
        S2GAE & 81.12$\pm$0.71 & 64.68$\pm$0.19 & 79.51$\pm$0.66 & 43.97$\pm$0.20 & 62.19$\pm$1.12 & 46.45$\pm$0.08 \\
        \bottomrule
    \end{tabular}}
\end{table}

\begin{table}[h!]
    \caption{The result of \texttt{GraphFM} in Link prediction with Node sampling. " - " means out of memory.}
    \label{lpnode}
    \centering
    \scalebox{0.85}{
    \begin{tabular}{cccccccc}
        \toprule
        Models & Metrics & Cora & Citeseer & Pubmed & Flickr & Reddit & Arxiv \\
        \midrule
        \multirow{2}{*}{BGRL} & 
        AUC & 98.53$\pm$1.07 & 99.41$\pm$0.61 & 99.60$\pm$0.04 & 87.88$\pm$0.24 & 42.72$\pm$3.84& 96.98$\pm$0.42 \\
        & AP & 98.75$\pm$0.74 & 99.53$\pm$0.38 & 99.52$\pm$0.05 & 88.24$\pm$0.18 & 59.22$\pm$0.95& 96.08$\pm$0.40 \\
        \multirow{2}{*}{CCA-SSG} & 
        AUC & 99.64$\pm$0.11 & 99.89$\pm$0.10 & 99.58$\pm$0.11 & 76.45$\pm$14.44 & 20.17$\pm$0.42& 45.28$\pm$0.52\\
        & AP & 99.63$\pm$0.14 & 99.87$\pm$0.12 & 99.49$\pm$0.16 & 73.96$\pm$17.05 & 45.62$\pm$0.72& 58.44$\pm$0.41\\
        \multirow{2}{*}{GBT} & 
        AUC & 99.22$\pm$0.10 & 99.74$\pm$0.10 & 99.40$\pm$0.10 & 86.69$\pm$0.42 & 46.86$\pm$0.44& 57.33$\pm$0.64\\
        & AP & 99.08$\pm$0.34 & 99.72$\pm$0.16 & 99.31$\pm$0.15 & 87.93$\pm$0.31 & 50.96$\pm$1.49& 60.61$\pm$0.70\\
        \multirow{2}{*}{GCA} & 
        AUC & 98.80$\pm$0.22 & 99.18$\pm$0.18 & 99.17$\pm$0.08 & \multirow{2}{*}{-} & \multirow{2}{*}{-} & \multirow{2}{*}{-} \\
        & AP & 98.58$\pm$0.30 & 99.11$\pm$0.26 & 99.08$\pm$0.12 & & & \\
        \multirow{2}{*}{GraphECL} & 
        AUC & 95.68$\pm$0.85 & 96.22$\pm$0.33 & 95.28$\pm$0.40 & \multirow{2}{*}{-} & \multirow{2}{*}{-} & \multirow{2}{*}{-} \\
        & AP & 95.44$\pm$1.15 & 96.61$\pm$0.38 & 94.67$\pm$0.48 & & & \\
        \midrule
        \multirow{2}{*}{GraphMAE} & 
        AUC & 97.28$\pm$0.29 & 99.33$\pm$0.43 & 96.04$\pm$0.09 & 50.00$\pm$0.00 & 72.49$\pm$0.43 & 50.01$\pm$0.02 \\
        & AP & 97.05$\pm$0.51 & 99.22$\pm$0.55 & 94.72$\pm$0.33 & 50.00$\pm$0.00 & 66.18$\pm$0.33 & 50.00$\pm$0.01 \\
        \multirow{2}{*}{GraphMAE2} & 
        AUC & 89.66$\pm$0.55 & 95.05$\pm$0.48 & 84.73$\pm$0.99 & 49.94$\pm$0.05 & 50.08$\pm$0.07 & 50.00$\pm$0.00 \\
        & AP & 90.80$\pm$0.91 & 95.08$\pm$0.11 & 85.37$\pm$0.85 & 49.98$\pm$0.03 & 50.04$\pm$0.03 & 50.00$\pm$0.00 \\
        \multirow{2}{*}{S2GAE} & 
        AUC & 95.16$\pm$0.30 & 95.75$\pm$0.12 & 88.77$\pm$0.64 & 49.95$\pm$0.44 & 90.75$\pm$1.12 & 94.97$\pm$0.69 \\
        & AP & 95.47$\pm$0.29 & 96.45$\pm$0.06 & 86.33$\pm$0.63 & 49.93$\pm$0.32 & 90.08$\pm$0.76 & 94.97$\pm$0.69 \\
        \bottomrule
    \end{tabular}}
\end{table}

\begin{table}[h!]
    \caption{The result of \texttt{GraphFM} in Link prediction with Subgraph sampling. " - " means out of memory.}
    \label{lpsub}
    \centering
    \scalebox{0.85}{
    \begin{tabular}{cccccccc}
        \toprule
        Models & Metrics & Cora & Citeseer & Pubmed & Flickr & Reddit & Arxiv \\
        \midrule
        \multirow{2}{*}{BGRL} &
        AUC & 97.18$\pm$1.96 & 99.81$\pm$0.09 & 99.47$\pm$0.07 & 86.92$\pm$0.15 & 21.99$\pm$0.23 & 93.81$\pm$3.20 \\
        & AP & 97.80$\pm$1.49 & 99.79$\pm$0.13 & 99.39$\pm$0.09 & 87.57$\pm$0.15 & 43.42$\pm$0.16 & 91.20$\pm$3.72 \\
        \multirow{2}{*}{CCA-SSG} &
        AUC & 99.91$\pm$0.00 & 98.23$\pm$0.29 & 99.69$\pm$0.00 & 50.00$\pm$0.00 & 21.22$\pm$0.08 & 33.25$\pm$0.64\\
        & AP & 99.91$\pm$0.00 & 97.85$\pm$0.24 & 99.62$\pm$0.04 & 50.00$\pm$0.00 & 45.88$\pm$0.54 & 49.22$\pm$2.56\\
        \multirow{2}{*}{GBT} &
        AUC & 99.13$\pm$0.07 & 99.83$\pm$0.11 & 98.70$\pm$0.14 & 50.00$\pm$0.00 & 44.72$\pm$0.07 & 45.70$\pm$2.80\\
        & AP & 99.03$\pm$0.09 & 99.75$\pm$0.23 & 98.83$\pm$0.05 & 50.00$\pm$0.00 & 48.01$\pm$0.33 & 50.45$\pm$2.79\\
        \multirow{2}{*}{GCA} &
        AUC & 98.61$\pm$0.21 & 99.97$\pm$0.03 & 98.85$\pm$0.25 & 50.00$\pm$0.00 &51.88$\pm$0.74 & 39.38$\pm$2.34\\
        & AP & 98.27$\pm$0.31 & 99.97$\pm$0.03 & 98.80$\pm$0.22 & 50.00$\pm$0.00 &50.97$\pm$0.82 & 50.10$\pm$1.43\\
        \multirow{2}{*}{GraphECL} &
        AUC & 98.98$\pm$0.19 & 99.65$\pm$0.27 & 96.23$\pm$0.14 & 51.06$\pm$0.35 & \multirow{2}{*}{-} & 59.01$\pm$0.54\\
        & AP & 98.62$\pm$0.60 & 99.64$\pm$0.28 & 95.69$\pm$0.31 & 51.12$\pm$0.41 & & 60.47$\pm$0.63\\
        \midrule
        \multirow{2}{*}{GraphMAE} &
        AUC & 96.63$\pm$0.79 & 99.65$\pm$0.20 & 95.86$\pm$0.25 & 50.00$\pm$0.00 & 77.07$\pm$1.42 & 91.50$\pm$0.56 \\
        & AP & 96.50$\pm$0.80 & 99.64$\pm$0.21 & 94.64$\pm$0.49 & 50.00$\pm$0.00 & 68.59$\pm$0.82 & 88.07$\pm$0.64 \\
        \multirow{2}{*}{GraphMAE2} &
        AUC & 89.62$\pm$0.54 & 94.92$\pm$0.57 & 83.99$\pm$1.02 & 49.94$\pm$0.05 & 50.00$\pm$0.00 & 99.01$\pm$0.00 \\
        & AP & 90.76$\pm$0.91 & 94.98$\pm$0.57 & 84.79$\pm$0.88 & 49.98$\pm$0.03 & 50.00$\pm$0.00 & 98.85$\pm$0.02 \\
        \multirow{2}{*}{S2GAE} &
        AUC & 95.92$\pm$1.28 & 96.76$\pm$0.49 & 89.72$\pm$0.37 & 49.91$\pm$0.23 & 80.17$\pm$0.93 & 79.00$\pm$1.31 \\
        & AP & 94.76$\pm$1.71 & 96.22$\pm$0.70 & 88.09$\pm$0.44 & 49.91$\pm$0.28 & 78.99$\pm$1.32 & 78.53$\pm$1.31 \\
        \bottomrule
    \end{tabular}}
\end{table}

\begin{table}[h!]
    \caption{The result of \texttt{GraphFM} in Node clustering with Node sampling. " - " means out of memory.}
    \label{clusteringnode}
    \centering
    \scalebox{0.8}{
    \begin{tabular}{ccccccc}
        \toprule
        Models & Cora & Citeseer & Pubmed & Flickr & Reddit & Arxiv \\
        \midrule
        BGRL & 0.3719/0.2217 & 0.1883/0.0628 & 0.1139/0.0588 & 0.0054/0.0145 & 0.5855/0.2043 & 0.2077/0.0472 \\
        CCA-SSG & 0.5107/0.4398 & 0.3525/0.2607 & 0.1246/0.0625 & 0.0622/0.0397 & 0.5560/0.1701 & 0.2868/0.0769\\
        GBT & 0.3441/0.1497 & 0.1466/0.0180 & 0.0453/-0.0084 & 0.0179/0.0080 & 0.5832/0.1680 & 0.2029/-0.0058\\
        GCA & 0.4591/0.3781 & 0.3229/0.2906 & 0.0708/0.0150 &- & -& -\\
        GraphECL & 0.4620/0.3262 & 0.3032/0.2123 & 0.1056/0.0199 & -& -& -\\
        \midrule
        GraphMAE & 0.5348/0.4476 & 0.3510/0.3042 & 0.1125/0.0250 & 0.0154/0.0197 & 0.8139/0.7313 & 0.3899/0.1848 \\
        GraphMAE2 & 0.3923/0.3286 & 0.2697/0.2714 & 0.2770/0.2594 & 0.0157/0.0097 & 0.6408/0.5023 & 0.3739/0.1747 \\
        S2GAE & 0.3457/0.2018 & 0.2179/0.1044 & 0.2647/0.2757 & 0.0067/0.0054 & 0.4464/0.2648 & 0.2800/0.1199 \\
        \bottomrule
    \end{tabular}}
\end{table}

\begin{table}[h!]
    \caption{The result of \texttt{GraphFM} in Node clustering with Subgraph sampling (NMI/ARI). " - " means out of memory. }
    \label{clusteringsub}
    \centering
    \scalebox{0.8}{
    \begin{tabular}{cccccccc}
        \toprule
        Models & Cora & Citeseer & Pubmed & Flickr & Reddit & Arxiv \\
        \midrule
        BGRL & 0.2589/0.1143 & 0.3102/0.1593 & 0.2272/0.1830 & 0.0052/0.0145 & 0.6227/0.1944 & 0.2123/0.0441 \\
        CCA-SSG & 0.2165/0.1258 & 0.1591/0.0265 & 0.2482/0.2214 & 0.0181/0.0125 & 0.5441/0.1764 & 0.2959/0.0846 \\
        GBT & 0.3657/0.1944 & 0.1373/0.0156 & 0.0638/-0.0018 & 0.0453/0.0176 & 0.6073/0.1610 & 0.2069/-0.0173 \\
        GCA & 0.4609/0.3277 & 0.2509/0.2125 & 0.0888/0.0182 & 0.0343/0.0155 & 0.5330/0.1709 &0.1732/0.0019 \\
        GraphECL & 0.5568/0.5186 & 0.3891/0.3663 & 0.3442/0.3057 & 0.0128/0.0142 & - & 0.3290/0.1296 \\
        \midrule
        GraphMAE & 0.5454/0.4424 & 0.4181/0.3975 & 0.3211/0.2839 & 0.0090/0.0137 & 0.7988/0.6419 & 0.4146/0.2035 \\
        GraphMAE2 & 0.3909/0.3271 & 0.2706/0.2695 & 0.2773/0.2598 & 0.0157/0.0068 & 0.4640/0.2320 & 0.2577/0.1149 \\
        S2GAE & 0.4121/0.2735 & 0.2762/0.2074 & 0.3001/0.2865 & 0.0039/0.0033& 0.3906/0.2112 & 0.2212/0.0666 \\
        \bottomrule
    \end{tabular}}
\end{table}

\subsection{Results of Efficiency}
\label{resultsofefficiency}
In Section 4.3, in addition to analyzing the performance of \texttt{GraphFM} under mini-batch conditions, we also examined the training efficiency of \texttt{GraphFM}. Further results are documented in %Figure~\ref{cora_mem},~\ref{citeseer_mem},~\ref{Flickr_mem},~\ref{Reddit_mem} and~\ref{arxiv_mem}, if the model point is not found on the figure, it indicates that the model got out-of-memory error during training. With the analysis detailed we can refer in Section 4.3.
table~\ref{efficiency} and ~\ref{efficiency_minibatch}, from the table, we can observe that compared to the other two training strategies, subgraph sampling requires less memory and provides faster training speeds. In terms of model comparison, S2GAE achieves better training efficiency across all benchmarked tasks.

\begin{table}[h!]
    \centering
    \caption{The memory usage of activations and the hardware throughput (higher is better).}
    \label{efficiency}
    \scalebox{0.85}{
    \begin{tabular}{cc|cc|cc|cc}
        \toprule
        & \multirow{3}{*}{Batch Type} & \multicolumn{2}{c|}{Cora} & \multicolumn{2}{c|}{Citeseer} & \multicolumn{2}{c}{Pubmed} \\
        \cmidrule{3-8}
        & & \multicolumn{1}{c}{Act Mem.} & \multicolumn{1}{c|}{Throughput} & \multicolumn{1}{c}{Act Mem.} & \multicolumn{1}{c|}{Throughput} & \multicolumn{1}{c}{Act Mem.} & \multicolumn{1}{c}{Throughput} \\
        & & \multicolumn{1}{c}{(MB)} & \multicolumn{1}{c|}{(iteration/s)} & \multicolumn{1}{c}{(MB)} & \multicolumn{1}{c|}{(iteration/s)} & \multicolumn{1}{c}{(MB)} & \multicolumn{1}{c}{(iteration/s)}\\
        \midrule
        \multirow{3}{*}{BGRL} & Full & 115.02 & 57.80 & 200.28 & 33.97 & 695.89 & 3.07 \\
        & Node & 101.48 & 51.02 & 126.90 & 34.68 & 411.47 & 28.17 \\
        & Subgraph & 24.55 & 85.48 & 24.14 & 91.17 & 22.75 & 86.34 \\
        \midrule
        \multirow{3}{*}{CCA-SSG} & Full & 123.70 & 4.36 & 263.72 & 2.24 & 552.28 & 2.39 \\
        & Node & 69.49 & 6.42 & 80.91 & 3.65 & 90.34 & 2.16 \\
        & Subgraph & 27.67 & 47.71 & 25.97 & 67.19 & 13.97 & 68.03 \\
        \midrule
        \multirow{3}{*}{GBT} & Full & 49.54 & 50.76 & 66.43 & 23.64 & 338.26 & 21.18 \\
        & Node & 65.04 & 58.82 & 76.14 & 51.02 & 93.29 & 64.51 \\
        & Subgraph & 95.04 & 54.64 & 174.47 & 28.73 & 82.43 & 37.59 \\
        \midrule
        \multirow{3}{*}{GCA} & Full & 430.84 & 17.94 & 354.45 & 9.40 & 8117.57 & 1.82 \\
        & Node & 364.88 & 6.01 & 351.15 & 3.00 & 3356.76 & 2.81 \\
        & Subgraph & 50.75 & 37.96 & 91.37 & 58.41 & 44.39 & 38.68 \\
        \midrule
        \multirow{3}{*}{GraphECL} & Full & 155.88 & 43.69 & 265.64 & 6.77 & 3554.12 & 1.72 \\
        & Node & 92.59 & 24.75 & 242.54 & 12.24 & 185.10 & 5.68 \\
        & Subgraph & 29.93 & 55.91 & 48.11 & 61.34 & 29.49 & 57.95 \\
        \midrule
        \midrule
        \multirow{3}{*}{GraphMAE} & Full & 142.08 & 48.08 & 370.15 & 29.88 & 577.61 & 27.24 \\
        & Node & 103.78 & 32.26 & 175.52 & 37.04 & 96.23 & 21.01 \\
        & Subgraph & 26.45 & 54.05 & 42.25 & 114.58 & 10.02 & 66.23 \\
        \midrule
        \multirow{3}{*}{GraphMAE2} & Full & 146.55 & 39.37 & 345.08 & 24.60 & 667.16 & 19.30 \\
        & Node & 57.17 & 37.04 & 86.15 & 23.20 & 63.95 & 22.52 \\
        & Subgraph & 18.31 & 48.77 & 44.92 & 102.30 & 13.61 & 60.24 \\
        \midrule
        \multirow{3}{*}{S2GAE} & Full & 28.09 & 26.18 & 35.43 & 1.38 & 205.03 & 2.30 \\
        & Node & 7.63 & 80.00 & 6.93 & 70.92 & 237.02 & 10.92 \\
        & Subgraph & 2.87 & 128.21 & 2.99 & 106.32 & 2.66 & 114.84 \\
        \bottomrule
    \end{tabular}}
\end{table}

% \begin{figure}[h!]
%     \centering
%     \includegraphics[scale=0.3]{cora_mem.pdf}
%     \caption{Time and space consumption of different methods and training strategy on Cora. }
%     \label{cora_mem}
% \end{figure}

% \begin{figure}[h!]
%     \centering
%     \includegraphics[scale=0.3]{citeseer_mem.pdf}
%     \caption{Time and space consumption of different methods and training strategy on Citeseer. }
%     \label{citeseer_mem}
% \end{figure}

% \begin{figure}[h!]
%     \centering
%     \includegraphics[scale=0.3]{Flickr_mem.pdf}
%     \caption{Time and space consumption of different methods and training strategy on Flickr. }
%     \label{Flickr_mem}
% \end{figure}

% \begin{figure}[h!]
%     \centering
%     \includegraphics[scale=0.3]{Reddit_mem.pdf}
%     \caption{Time and space consumption of different methods and training strategy on Reddit. }
%     \label{Reddit_mem}
% \end{figure}

% \begin{figure}[h!]
%     \centering
%     \includegraphics[scale=0.3]{arxiv_mem.pdf}
%     \caption{Time and space consumption of different methods and training strategy on Ogbn-arxiv. }
%     \label{arxiv_mem}
% \end{figure}

\begin{table}[h!]
    \centering
    \caption{The memory usage of activations and the hardware throughput. " - " means out of memory.}
    \label{efficiency_minibatch}
    \scalebox{0.87}{
    \begin{tabular}{cc|cc|cc|cc}
        \toprule
        & \multirow{3}{*}{Batch Type} & \multicolumn{2}{c|}{Flickr} & \multicolumn{2}{c|}{Reddit} & \multicolumn{2}{c}{Arxiv} \\
        \cmidrule{3-8}
        & & \multicolumn{1}{c}{Act Mem.} & \multicolumn{1}{c|}{Throughput} & \multicolumn{1}{c}{Act Mem.} & \multicolumn{1}{c|}{Throughput} & \multicolumn{1}{c}{Act Mem.} & \multicolumn{1}{c}{Throughput} \\
        & & \multicolumn{1}{c}{(MB)} & \multicolumn{1}{c|}{(iteration/s)} & \multicolumn{1}{c}{(MB)} & \multicolumn{1}{c|}{(iteration/s)} & \multicolumn{1}{c}{(MB)} & \multicolumn{1}{c}{(iteration/s)}\\
        \midrule
        \multirow{2}{*}{BGRL}
        & Node & 1379.81 & 6.64 & 2490.87 & 2.56 & 2476.28 & 4.47 \\
        & Subgraph & 22.61 & 126.58 & 21.63 & 89.29 & 87.26 & 37.73 \\
        \midrule
        \multirow{2}{*}{CCA-SSG}
        & Node & 612.13 & 2.27 & 1496.45 & 0.81 & 2841.26 & 1.12 \\
        & Subgraph & 13.90 & 58.82 & 15.05 & 42.74 & 31.84 & 0.001 \\
        \midrule
        \multirow{2}{*}{GBT}
        & Node & 1801.95 & 5.14 & 4335.30 & 1.74 & 3030.22 & 1.72 \\
        & Subgraph & 219.11 & 37.87 & 271.24 & 8.25 & 224.10 & 24.75 \\
        \midrule
        \multirow{2}{*}{GCA}
        & Node & - & - & - & - &-  & - \\
        & Subgraph & 44.80 & 35.34 & 36.82 & 4.75 & 264.96 & 0.79 \\
        \midrule
        \multirow{2}{*}{GraphECL}
        & Node & - & - & - & - & - & - \\
        & Subgraph & 22.14 & 34.97 & - & - & 20.16 & 19.65 \\
        \midrule
        \midrule
        \multirow{2}{*}{GraphMAE}
        & Node & 638.83 & 5.54 & 1293.80 & 1.57 & 399.42 & 6.71 \\
        & Subgraph & 10.36 & 63.69 & 12.38 & 47.39 & 6.49 & 21.50 \\
        \midrule
        \multirow{2}{*}{GraphMAE2}
        & Node & 1039.38 & 3.29 & 2525.84 & 1.63 & 906.31 & 6.24 \\
        & Subgraph & 16.29 & 48.54 & 22.80 & 46.73 & 14.14 & 22.27 \\
        \midrule
        \multirow{2}{*}{S2GAE}
        & Node & 165.50 & 17.15 & 380.07 & 1.42 & 179.94 & 2.92 \\
        & Subgraph & 2.62 & 135.14 & 0.87 & 136.99 & 2.61 & 43.29 \\
        \bottomrule
    \end{tabular}}
\end{table}

\subsection{Results of Early Stop Criteria}
\label{stopcri}
In Section 4.4, we analyzed the performance of \texttt{GraphFM} when different downstream tasks were used as early stopping criteria. More experimental results are documented in Tables~\ref{clusteringsavemodel_flickr},~\ref{lpsavemodel_arxiv} and~\ref{clustersavemodel_arxiv}. Due to tests on the Reddit dataset typically taking more than 24 hours, this study primarily conducts tests on the Flickr and Ogbn-arxiv datasets. We use AUC in link prediction and NMI in node clustering as the metrics to preserve the valid model, and the analysis is detailed in Section 4.4. 

\begin{table}[h!]
    \caption{The result of \texttt{GraphFM} in Flickr dataset by saving valid model with the best performance in node clustering. " - " means out of memory.}
    \label{clusteringsavemodel_flickr}
    \centering
    \begin{tabular}{ccccc}
        \toprule
        Training Strategy& Models & Node Classification & Link Prediction & Node Clustering \\
        \midrule
        \multirow{8}{*}{Node Sampling}&BGRL & 46.33$\pm$0.94 & 86.56/87.33 & 0.0094/0.0181  \\
        &CCA-SSG & 51.75$\pm$1.22 & 90.08/90.56 & 0.0850/0.0583  \\
        &GBT & 51.70$\pm$0.86 & 97.21/97.39 & 0.0453/0.0196 \\
        &GCA & - & - &  - \\
        &GraphECL & - & - &  - \\
        \cmidrule(r){2-5}
        &GraphMAE & 46.68$\pm$0.74 & 50.00/50.00 & 0.0296/0.0282 \\
        &GraphMAE2 & 46.54$\pm$0.68 & 50.00/50.00 & 0.0266/0.0081 \\
        &S2GAE & 45.13$\pm$0.32 & 50.65/50.41 & 0.0246/0.0175 \\
        \midrule
        \multirow{8}{*}{Subgraph Sampling}&BGRL & 46.32$\pm$0.78 & 86.86/87.52 &  0.0082/0.0183 \\
        &CCA-SSG & 49.51$\pm$1.03 & 55.39/52.86 & 0.0986/0.0579  \\
        &GBT & 52.36$\pm$0.32 & 91.08/91.70 & 0.0699/0.0328 \\
        &GCA & 49.87$\pm$0.88 & 55.45/53.18 &  0.0887/0.0393  \\
        &GraphECL & 46.42$\pm$1.20 & 60.70/60.39 &  0.0526/0.0466 \\
        \cmidrule(r){2-5}
        &GraphMAE & 46.46$\pm$0.94 & 50.00/50.00 & 0.0261/0.0099 \\
        &GraphMAE2 & 45.86$\pm$0.09 & 50.10/50.05 & 0.0249/0.0109 \\
        &S2GAE & 44.76$\pm$0.43 & 50.25/50.30 & 0.0068/0.0049 \\
        \bottomrule
    \end{tabular}
\end{table}

\begin{table}[h!]
    \caption{The result of \texttt{GraphFM} in ogbn-arxiv dataset by saving valid model with the best performance in link prediction. " - " means out of memory.}
    \label{lpsavemodel_arxiv}
    \centering
    \begin{tabular}{ccccc}
        \toprule
        Training Strategy& Models & Node Classification & Link Prediction & Node Clustering \\
        \midrule
        \multirow{8}{*}{Node Sampling}&BGRL & 64.97$\pm$0.61 & 98.98/98.95 & 0.2058/0.0415  \\
        &CCA-SSG & 67.08$\pm$0.43 & 99.33/99.30 & 0.2801/0.0429  \\
        &GBT & 62.70$\pm$1.07 & 98.78/98.69 & 0.2220/-0.0166 \\
        &GCA & - & - &  - \\
        &GraphECL & - & - &  - \\
        \cmidrule(r){2-5}
        &GraphMAE & 65.64$\pm$0.93 & 90.85/87.85 & 0.3971/0.1851 \\
        &GraphMAE2 & 64.82$\pm$0.72 & 50.09/50.04 & 0.3760/0.1873 \\
        &S2GAE & 43.65$\pm$2.11 & 80.50/78.61 & 0.2064/0.0708 \\
        \midrule
        \multirow{8}{*}{Subgraph Sampling}&BGRL & 64.63$\pm$0.87 & 99.11/99.09 &  0.2198/0.0549 \\
        &CCA-SSG & 61.90$\pm$0.73 & 97.58/97.60 & 0.2697/0.0476  \\
        &GBT & 56.44$\pm$2.13 & 87.57/80.41 & 0.1300/-0.0116 \\
        &GCA & 59.99$\pm$0.82 & 96.00/95.46 & 0.2496/0.0070  \\
        &GraphECL & 55.42$\pm$0.95 & 93.38/93.16 &  0.3114/0.1375 \\
        \cmidrule(r){2-5}
        &GraphMAE & 66.43$\pm$0.53 & 89.70/86.17 & 0.4146/0.1985 \\
        &GraphMAE2 & 64.42$\pm$1.01 & 71.96/64.71 & 0.3732/0.1763 \\
        &S2GAE & 38.06$\pm$2.43 & 77.50/76.84 & 0.1663/0.0426 \\
        \bottomrule
    \end{tabular}
\end{table}

\begin{table}[h!]
    \caption{The result of \texttt{GraphFM} in ogbn-arxiv dataset by saving valid model with the best performance in node clustering. " - " means out of memory.}
    \label{clustersavemodel_arxiv}
    \centering
    \begin{tabular}{ccccc}
        \toprule
        Training Strategy& Models & Node Classification & Link Prediction & Node Clustering \\
        \midrule
        \multirow{8}{*}{Node Sampling}&BGRL & 64.64$\pm$0.43 & 98.68/98.57 & 0.2384/0.0605  \\
        &CCA-SSG & 43.69$\pm$0.73 & 94.32/92.15 & 0.2381/0.0305  \\
        &GBT & 59.40$\pm$0.71 & 84.57/78.01 & 0.2130/-0.0168 \\
        &GCA & - & - &  - \\
        &GraphECL & - & - &  - \\
        \cmidrule(r){2-5}
        &GraphMAE & 68.56$\pm$0.94 & 88.71/85.40 & 0.4138/0.1976 \\
        &GraphMAE2 & 67.58$\pm$0.08 & 50.00/50.00 & 0.3950/0.1871 \\
        &S2GAE & 52.19$\pm$1.45 & 86.83/85.92 & 0.3126/0.1237\\
        \midrule
        \multirow{8}{*}{Subgraph Sampling}&BGRL & 64.17$\pm$0.46 & 98.39/98.21 &  0.2268/0.0572 \\
        &CCA-SSG & 49.97$\pm$0.62 & 93.89/93.93 & 0.3358/0.0787  \\
        &GBT & 59.85$\pm$1.34 & 50.00/50.00 & 0.1903/-0.0146 \\
        &GCA & 61.64$\pm$0.95 & 93.76/89.32 & 0.2696/-0.0002 \\
        &GraphECL & 54.04$\pm$1.42 & 92.97/92.70 &  0.3003/0.1406 \\
        \cmidrule(r){2-5}
        &GraphMAE & 68.13$\pm$0.53 & 91.31/87.71 & 0.4055/0.1861 \\
        &GraphMAE2 & 67.57$\pm$0.87 & 50.00/50.00 & 0.3919/0.1783 \\
        &S2GAE & 46.50$\pm$1.16 & 83.01/81.88 & 0.2538/0.0805 \\
        \bottomrule
    \end{tabular}
\end{table}

\end{document}